\documentclass[sn-basic]{sn-jnl}

\newcommand{\comment}[1]{}
\usepackage{lscape}
\usepackage{adjustbox}
\usepackage{soul}
\usepackage{caption}
\usepackage{float}
\usepackage{graphicx}
\usepackage{rotating}  

\jyear{2022}%

\theoremstyle{thmstyleone}%
\newtheorem{theorem}{Theorem}
\newtheorem{proposition}[theorem]{Proposition}%

\theoremstyle{thmstyletwo}%
\newtheorem{example}{Example}%
\newtheorem{remark}{Remark}%

\theoremstyle{thmstylethree}%
\newtheorem{definition}{Definition}%

\begin{document}


\title[A Novel Approach for Intrinsic Dimension Estimation...]{A Novel Approach for Intrinsic Dimension Estimation via Ritz Values and Orthogonal Polynomials}


\author*[1]{\fnm{Kadir} \sur{Özçoban}}\email{kadir.ozcoban@metu.edu.tr}

\author[1]{\fnm{Murat} \sur{Manguoğlu}}\email{manguoglu@ceng.metu.edu.tr}

\author[2]{\fnm{E. Fatih} \sur{Yetkin}}\email{fatih.yetkin@khas.edu.tr}

\affil*[1]{\orgdiv{Department of Computer Engineering}, \orgname{Middle East Technical University}, \orgaddress{ \city{Ankara}, \postcode{06800}, \country{Turkey}}}

\affil[2]{\orgdiv{Management Information Systems}, \orgname{Kadir Has University}, \orgaddress{\city{Istanbul}, \postcode{34083}, \country{Turkey}}}


\abstract{
The real-life data have a complex and non-linear structure due to their nature. These non-linearities and  
large number of features can usually cause problems such as the empty-space phenomenon and the well-known curse of dimensionality. Finding the nearly optimal representation of the dataset in a lower-dimensional space (i.e. dimensionality reduction) offers an applicable mechanism for improving the success of machine learning tasks.  
However, estimating the required data dimension for the nearly optimal representation (intrinsic dimension) can be very costly, particularly if one deals with big data. We propose a highly efficient and robust intrinsic dimension estimation approach that only relies on matrix-vector products for dimensionality reduction methods. An experimental study is also conducted to compare the performance of proposed method with state of the art approaches. }




\maketitle




\maketitle

\section{Introduction}\label{sec1}
Dimensionality reduction approaches are crucial 
in various applications of machine learning tasks 
such as computer vision, robotics, natural language processing, medical diagnosis, recommendation systems or industrial IoT applications such as predictive maintenance which need to generate and process large amounts of data and variables. 
In general, dimensionality reduction improves the performance  of machine learning tasks' by removing redundant features. In this regard, both linear and non-linear dimensionality reduction methods, specifically the manifold learning techniques are particularly efficient since they are based on the preservation of the geometric structure of the original feature space. In this manner, there are several approaches  already available and studied extensively in the literature such as principal component analysis (PCA), Multidimensional scaling (MDS), Laplacian Eigenmaps (LE) and other. We refer the reader to \citep{survey} for a comprehensive survey of the available methods. However, it is well-known that these methods have two computational bottlenecks: i)construction of the variance (PCA), distance (MDS), or neighborhood (graph-embedding based techniques such as LE) matrices; ii)calculation of the solution of ordinary or generalized eigenvalue problems. 

Due to these two bottlenecks, most of the dimensionality reduction methods are not scalable in terms of the problem size due to the rapid explosion of data at several domains and therefore they are not applicable for big data problems in a plain form \citep{megaman}. Nevertheless, there are variety of 
approaches already available in the literature to speed-up the methods especially in high-performance computing platforms \citep{paralelEfforts}. Another way to improve the efficiency of these methods is to use a predetermined intrinsic dimension while solving the eigenvalue problem arising from the optimization problem that occurs during the dimensionality reduction. In the literature, one can find well-established, very efficient and accurate eigensolvers \citep{slepc,feast,sameh1982trace}. Especially Krylov subspace based eigensolvers offer a fast convergence  rate 
for finding the eigenpairs in a given interval. Intrinsic dimension estimators can exploit the advantages of Krylov subspace based methods, as proposed in this paper, to create an efficient approach for the determination of a nearly-optimal reduction order. 
  
We propose a novel approach for estimating the intrinsic dimension of a given dataset. We employed several numerical techniques in the proposed method to avoid the expensive eigenvalue computations, which is the main bottleneck of most well-known intrinsic dimension estimation methods.  In the proposed method, we applied several different algorithms in coherence, such as Conjugate Gradient and trace estimation. The main novelty of this approach is to construct the estimation of the intrinsic dimensionality problem  only via matrix-vector products. In that way, the problem can easily be handled in large dimensions at high-performance computing systems since many existing state-of-the-art algorithms exist for the parallel matrix-vector multiplications. 


The rest of the paper is organized as follows. In Section 2, we will briefly discuss the existing methods. In Section 3, we will present the mathematical tools we have employed to develop the proposed method and the details of the proposed method. We demonstrated the numerical results for a toy and artificially generated large data in Section 4. Finally, in Section 5, we present the conclusions.

\section{Related Work}\label{sec2}

The existing intrinsic dimension (ID) estimators 
can be grouped into three categories: i) topological, ii) graph-based, and iii) projective methods~\citep{bib1}. These methods have several advantages (such as high accurate estimation capability of topological estimators) discussed in this section. However, projective methods can be considered better in terms of computational complexity if an efficient eigensolver is implemented. In this paper, we propose an approach to improve its computational complexity by avoiding the eigen-decomposition. Since our proposed method is related to projective methods, they will in detail as well. We will also briefly introduce the other techniques for comparison. 
\subsection{\textit{Topological Estimators}}
Topological estimators assume that the data has a shape of a lower-dimensional manifold $\mathcal{M} \in \mathcal{R}^d$ embedded in a higher dimensional space $\mathcal{R}^D$ through a locally smooth map $\phi:\mathcal{M} \rightarrow \mathcal{R}^D$  and accept the ID of the data as the topological dimension of the manifold.\\ \\
\textbf{Definition 1}\textit{(open cover)} Given a topological space $\mathcal{X}$ and a subset $\mathcal{Y}\subseteq \mathcal{X}$, a collection of open sets $\mathcal{U}=\cup_{i\in I}U_i$  is called an open cover of $\mathcal{Y}$ if $\mathcal{Y}\subseteq \mathcal{U}$ where $I$ and $U_i$ are an index set and an open set respectively.\\ \\
\textbf{Definition 2}\textit{(refinement of a cover)} A cover $\mathcal{U}'=\cup_{i\in I} U_i'$ is a refinement of a cover $\mathcal{U}=\cup_{i\in I}U_i$ if $\forall U'_{\in \mathcal{U'}} \exists U_{\in \mathcal{U}}$ such that $U' \subseteq U$.\\ \\
\textbf{Definition 3}\textit{(Topological Dimension as known as Lebesgue Covering
Dimension)} The topological dimension of a topological space $\mathcal{X}$ is the smallest integer \textit{d} if every finite covering $\mathcal{U}$ of $\mathcal{X}$ has a refinement $\mathcal{U}'$ such that no subset of $\mathcal{X}$ has more than \textit{d}+1 intersecting open sets in $\mathcal{U}'$. If no minimal integer exists, $\mathcal{X}$ is said to have infinite topological
dimension. \\ 

In~\citep{bib2}, a method for estimating the ID of the data is proposed by assuming that the data has the shape of a manifold in a lower dimension and the ID is accepted as the topological dimension of that manifold. However, as the authors pointed out,  the proposed approach 
requires that the data set is well-sampled on a smooth manifold. Therefore, is not applicable for a noisy data set. 

Since the topological dimension is practically inestimable, other alternatives have also been proposed. They can be categorized under two headings, \textit{fractal}, and \textit{nearest-neighbors-based} id estimators. 
\subsubsection{\textit{Fractal Estimators}}
In fractal ID estimators, it is assumed that $\mathcal{M}$ has somehow fractal structure, and ID is estimated by employing fractal dimension estimation. They are based on the idea that the volume of $d$ dimensional set scales with its size $r$ as $r^d$. So to estimate the fractal dimension, the number of neighborhoods in a radius $r$ is counted, and its rate of growth based on $r$ is computed. For $d$ dimensional fractal, the rate of growth is $r^d$. Correlation Dimension (CD) Estimation~\citep{bib21} is a well-known example for  \textit{fractal-based} ID estimators.  
\\ \\
\textbf{Definition 4}~\textit{(Correlation Dimension)} \\
Let $X_N$ be a finite sample set, $\|\cdot\|$ be the Euclidean norm and $I(\cdot)$ be the step function with the property $I(y) = 0 $ if y$<$0, and $I(y)=1$ otherwise. Then, \\
\begin{equation*}
    C(n,r) = \frac{2}{n(n-1)}\sum_{i=1}^n\sum_{j=i+1}^nI(r-\|x_i-x_j\|)
\end{equation*}
 is defined as the correlation integral. It computes the probability of the distance between two points randomly selected from the set being less than or equal to $r$. And CD is given as   
 \begin{equation*}
     \lim_{r \to 0}\lim_{n \to \infty}\frac{\text{log}~C(n,r)}{\text{log}~r}.
 \end{equation*}

In practical implementations, CD estimation finds an ID approximation by evaluating $C(n,r)$ with different $r_i$ values and 
solving a linear least squares problem to the 
computed data points ($\text{log}~r_i$, $\text{log}~C(n,r_i)$). It should be noted that for a correct estimation, the number of data points must be very large~\citep{bib22}. 
\subsubsection{\textit{Nearest-Neighbors-Based Estimators}}
\textit{Nearest-Neighbors-Based ID Estimators} approximate the ID by the distributions of data neighbors. One of the most cited example is Maximum Likelihood Estimator (MLE) which uses Poisson process approximation for the ID estimation and has $O(N^2D)$ time complexity~\citep{bib23}. It estimates the local intrinsic dimension of $x_i$ as
\begin{equation*}
    \hat{d}(x_i,k) = \Biggl(\frac{1}{k}\sum_{j=1}^k\text{log}\frac{T_{k+1}(x_i)}{T_j(x_i)}\Biggr)^{-1}
\end{equation*}
where $T_j(x_i)$ is the distance between $x_i$ and its $j^{th}$ nearest neighbor.
To estimate global ID, the average of $\hat{d}(x_i,k)$ values for all data points should be considered as $\hat{d}(k)=\frac{1}{N}\sum_{i=1}^N\hat{d}(x_i,k)$.

A recent ID estimator based on MLE techniques is TLE~\citep{bib30}. Based on a new extreme-value theoretic analysis, the estimator employs MLE methods over all known pairwise distances among the sample participants. Also, an ID estimator recently proposed based on Methods of Moments(MOM)~\citep{bib31}.

Another proposed \textit{Nearest-Neighbors-Based ID Estimator} is Manifold-Adaptive Dimension Estimation (MADA)~\citep{bib29} in which first the local ID's around data points are estimated by nearest neighbors techniques, and these are combined.

 ESS ID estimator which estimates the ID based on the expected simplex skewness measure having $O(N^{d+1}D(d+1)^2)$ time complexity is another type of \textit{Nearest-Neighbors-Based Estimators}~\citep{bib32}.

A family of ID estimators using a similar approach is the Minimum Neighbor Distance—Maximum Likelihood ($\text{MiND}_{\text{ML*}}$) estimators which utilize a maximum likelihood method on the probability density function pertaining to the normalized nearest neighbor distances~\citep{bib24}\comment{employs the probability distribution function 
of the normalized nearest neighbor distances~\citep{bib24}.(direkt paperdan aldım bu cümleyi, değiştirmek gerekebilir)}. Let $p(x_i)$ be the ratio of the distance between $x_i$ and its nearest neighbor and the distance between $x_i$ and its $(k+1)^{th}$ nearest neighbor then $ll(d)$ can be defined as
\begin{equation*}
    ll(d) = \sum_{x_i \in X_N} \text{log}g(x_i;k,d)
\end{equation*}
\begin{equation*}
    = N\text{log}k + N\text{log}d + (d-1)\sum_{x_i \in X_N} \text{log}p(x_i) + (k-1) \sum_{x_i \in X_N} \text{log}(1-p^d(x_i)). 
\end{equation*}
In $\text{MiND}_{\text{MLi}}$(integer variant of $\text{MiND}_{\text{ML*}}$), ID is estimated by $\hat{d} = \underset{d \in \{1..D\}}{\text{argmax}} ll(d)$ with time complexity $O(N^2D^2)$~\citep{bib24}. 

For a real number ID approximation, they also propose $\text{MiND}_{\text{MLk}}$ which finds the maximum value of $ll(d)$ in $[1,D]$. It looks for the solution of $\frac{\partial ll}{\partial d}=0$,
\begin{equation*}
    \frac{N}{d} + \sum_{x_i \in X_N}\biggl(\text{log}p(x_i)-(k-1)\frac{p^d(x_i)\text{log}p(x_i)}{1-p^d(x_i)}\biggr) = 0
\end{equation*}
and 
\begin{equation*}
    \hat{d} = \underset{0<d \leq D}{\text{argmax }} ll(d).
\end{equation*} 
The time complexity of $\text{MiND}_{\text{MLk}}$ is $O(N^2D^2)$~\citep{bib24} and it should be noted that for $k=1$ it is same as MLE. 

Also in the same study, they propose $\text{MiND}_{\text{KL}}$ estimator which also has $O(N^2D^2)$ time complexity and estimates ID as
\begin{equation*}
   \hat{d} = \underset{d\in \{1..D\}}{\text{argmin}}\biggl(\text{log}\frac{N}{N-1}+\frac{1}{N}\sum_{i=1}^N\text{log}\frac{\check{p}(\hat{r}_i)}{\hat{p}_d(\hat{r}_i)}\biggr)
\end{equation*}
where  $\check{p}_d(\hat{r}_i)$ are the distances between $\hat{r}_i$ and its first neighbor in $\hat{r}$ and in $\check{r}_d$, respectively, in which $\hat{r} = \{\hat{r}_i\}_{i=1}^N=\{p(x_i)\}_{i=1}^N$, $\hat{p}(r)=\frac{N^{-1}}{2p(r)}$ $p(r)$ is the distance between r and its nearest neighbor. 

A more advanced ID estimator DANCo~\citep{bib27} has been proposed which is motivated by $\text{MiND}_{\text{KL}}$ and has $O(N\text{log}(N)D^2)$ time complexity. DANCo exploits angle
and norm concentrations and estimates ID as
\begin{equation*}
    \hat{d} = \underset{d\in \{1..D\}}{\text{argmin}} \mathcal{K}\mathcal{L}(g_{Data},g_{Sphere}^d) +  \mathcal{K}\mathcal{L}(q_{Data},q_{Sphere}^d)
\end{equation*}
where $\mathcal{K}\mathcal{L}$ is Kullback–Leibler divergence operator, $g(r;k,d)=kdr^{d-1}(1-r^d)^{k-1}$, and $q(x;v,t)$ is the von Mises-Fisher distribution. Also in the same study, authors propose FastDANCo ID estimator which has $O(N\text{log}(N)D)$ time complexity by performing some relaxations on DANCo~\citep{bib27}.

A recently proposed ID estimator is TwoNN which  computes the ID based on only two nearest neighbors of the data points~\citep{bib26}. Let $\mu$ be the ratio of the distance between $x_i$ and its second nearest neighbor and the distance between $x_i$ and its nearest neighbor and $F(\mu)=(1-\mu^{-d})1_{[1,+\infty]}(\mu)$, then ID can be estimated as
\begin{equation*}
   \hat{d}=\frac{\text{log}(1-F(\mu))}{\text{log}(\mu)}.
\end{equation*}


\subsection{\textit{Graph-Based Estimators}}

Graph-based ID estimation approaches have some advantages, as they usually have lower computational complexity and due to the availability of theoretical tools in literature~\citep{bib3}.


Examples of such estimators rely on different graph-theoretic structures such as, the ID estimators proposed in \citep{bib3} and \citep{bib4} are based on \textit{k}-sphere of influence graph and minimum spanning tree respectively.  

In \citep{bib25}, the authors proposed an estimator based on nearest neighbors information. Hereinafter, it will be referred to as kNNG. In which $kNNG(X_N)$ is k-nn graph of $X_N$, $MST(X_N)$ is the minimum spanning tree of $kNNG(X_N)$, $\mathcal{L}(MST(X_N))= \sum\|e_{i,j}\|^\gamma, \gamma \in (0,d)$, $\text{log}(\mathcal{L}(MST(X_N))=a\text{log(d)+b},a=\frac{d-\gamma}{d}$. To estimate d a set of cardinalities $\{n_k\}_{k=1}^K$ is chosen, and set of $(\text{log}\mathcal{L}(MST(X_{n_k})),n_k)$ is computed. Then, the least squares method is performed on that set to find $\hat{a} \approx$ a. Finally, ID is approximated as round$\{\gamma/(1-\hat{a})\}$.

\subsection{\textit{Projective Estimators}}
Projective ID estimation methods search for the best subspace which has minimum projection error to project the data~\citep{bib5}. Here it should be noted that, their initial design purpose was dimensionality reduction. However, with some modifications, they can also be used for ID estimation. Most of them can be categorized into two types of techniques, Multidimensional Scaling(MDS) and PCA \citep{bib1}.

MDS methods are projection techniques that have the objective of preserving the pairwise distances of the data~\citep{bib6}. Some known examples are Bennett’s algorithm~\citep{bib7}, Sammon’s
mapping~\citep{bib8}, Curvilinear Component Analysis (CCA)~\citep{bib9}, Local Linear Embedding (LLE)~\citep{bib10}.

Due to the its variance-oriented reduction mechanism, PCA~\citep{bib11} can be considered as a very favorable technique for the ID estimation. It aims to re-express the data with new bases, a combination of original ones, in a way that re-expressed data has maximum variance and minimum covariance. To perform that re-expression, data matrix $X \in \mathcal{R}^{NxD}$ is multiplied with orthonormal \textit{principal component matrix} $\mathcal{P}$ whose columns are named as principal components. And these principal components are the eigenvectors of the $\textit{covariance matrix }$ $\mathcal{C}=\frac{1}{N-1}X_{C}^TX_{C}$ where $X_{C}$ is the centered version of $X$. Namely, $X_{C}=X-\text{mean}(X)$.

In PCA, associated eigenvalues of eigenvectors give some information about the importance of the eigenvector in the basis change process. Namely, the greater the eigenvalue, the more information of the data in the direction of the associated eigenvector. Therefore, eigenvalues of the $\textit{covariance matrix}$ are used to estimate ID using PCA. The main idea is to eliminate relatively less important eigenvectors and keep the dominant ones. To do that a threshold $0 < \theta <1$ or $\alpha \gg $1 is determined, and ID is approximated as either based on the ratio of total variance:
\begin{equation*}
    d = \underset{k}{\text{argmin}}  \left\{ \frac{\sum_{i=1}^k \lambda_i}{\sum_{i=1}^D \lambda_i} \geq \theta \right\}
\end{equation*}
or either based on the ratio between consecutive variances:
\begin{equation*}
    d = \underset{k}{\text{argmin}}  \left\{ \frac{\lambda_k}{\lambda_{k+1}} \geq \alpha  \right\}
\end{equation*}
where $\lambda_1\ge\lambda_2\ge\hdots\ge\lambda_D\ge0$ are the eigenvalues of $\mathcal{C}$.

PCA works well for linear data; different approaches have been proposed to extend it to non-linear data. One of them is  Local PCA (LPCA), which suggests that firstly data should be separated into sub-regions, later PCA should be applied to each region separately. Then, the ID of the data can be determined by combining  estimated sub-region's IDs~\citep{bib13}.

Based on the Local PCA idea, another PCA-based ID estimator has been proposed in~\citep{bib28}, in which  an approximate minimal set cover algorithm and  PCA applied locally to each subset of the cover. Also, in the same study, they proposed a noise filtering approach for PCA.

Another proposed alternative for nonlinear extension is called Kernel PCA (KPCA)~\citep{bib12}. KPCA tries to adapt PCA to a non-linear form by using a kernel function in the construction of covariance matrix.

One of the recent\comment{and accurate} ID estimators in literature is the Fisher Separability algorithm (FSH)~\citep{bib19}, which estimates the ID of the data based on separability properties. At the beginning of the algorithm, the data is projected into the linear subspace spanned by the first $k$ principal components where k is determined by $k=max\{i:\lambda_1/\lambda_i < C\}$ with $C$ being a constant. 

\comment{
At the beginning of the algorithm, the data is projected into the linear subspace spanned by the first $k$ principal components. The number k is determined by $k=max\{i:\lambda_D/\lambda_i < C\}$ where C is constant (default 10). Since $k$ is not known beforehand, all principal components are computed. That is a very costly operation, especially for big data. In our proposed technique, we estimate the number of eigenvalues in an interval for the covariance matrix without constructing it explicitly. After determining the $k$ value, only the $k$ principal components can be efficiently computed.

}

\section{Methodology}\label{sec3}
In PCA based methods, two essential steps are the construction of a covariance matrix and its eigen-decomposition. These two steps become more and more costly as the dimension of the data increases. Therefore, in our proposed method, we aim to estimate the ID of the data by avoiding these steps. Instead of computing the eigen-decomposition of a covariance matrix, we approximate the eigenvalue distribution of it without constructing it. 

The proposed approach can be divided into three main building blocks: a) estimation of the trace of the covariance matrix, which is equal to the summation of its eigenvalues, b) construction of the eigenvalue search intervals, c) estimation of the number of eigenvalues in the intervals to approximate the distribution of eigenvalues and estimate ID. 

\subsection{Trace Estimation of Covariance Matrix}
For a \textit{symmetric positive semi-definite} matrix $\mathcal{A} \in \mathcal{R}^{nxn}$, and the vector $z$ whose 
components are \textit{Rademacher random variables}, equal to $+$1 or $-$1 with
 equal probability, the trace can computed as~\citep{bib14}
\begin{equation}
\label{eqn:trace}
    tr(\mathcal{A}) = E[z^T\mathcal{A}z],~z\sim ^{iid} \{-1,1\}^ n
\end{equation}
where idd is the abbreviation for Independent and Identically Distributed.

Hence, the trace can be approximated by averaging over some random samples
\begin{equation*}
    tr(\mathcal{A}) \approx \frac{1}{n_v}\sum_{k=1}^{n_v}z_k^T\mathcal{A}z_k.
\end{equation*}
But we have the data matrix $X \in \mathcal{R}^{NxD}$ and its centered version $X_C$, and require the trace of $\mathcal{C}=\frac{1}{N-1}X_{C}^TX_{C}$.
\begin{eqnarray}
    tr(\mathcal{C}) &\approx& \frac{1}{n_v}\sum_{k=1}^{n_v}z_k^T\mathcal{C}z_k \nonumber \\ 
    tr(\mathcal{C}) &\approx& \frac{1}{n_v}\sum_{k=1}^{n_v}z_k^T\frac{X_{C}^TX_{C}}{N-1}z_k \nonumber \\
    tr(\mathcal{C}) &\approx& \frac{1}{N-1}\biggl(\frac{1}{n_v}\sum_{k=1}^{n_v}z_k^TX_{C}^TX_{C}z_k\biggr) \nonumber \\
    tr(\mathcal{C}) &\approx& \frac{1}{N-1}\biggl(\frac{1}{n_v}\sum_{k=1}^{n_v}(X_{C}z_k)^TX_{C}z_k \biggr)\nonumber \\
    tr(\mathcal{C}) &\approx& \frac{1}{N-1}\biggl(\frac{1}{n_v}\sum_{k=1}^{n_v}h_k^Th_k\biggr)
\end{eqnarray}
where $h_k = X_{C}z_k$. 

\begin{figure}[h!]
\centering
  \includegraphics[width=0.8\textwidth]{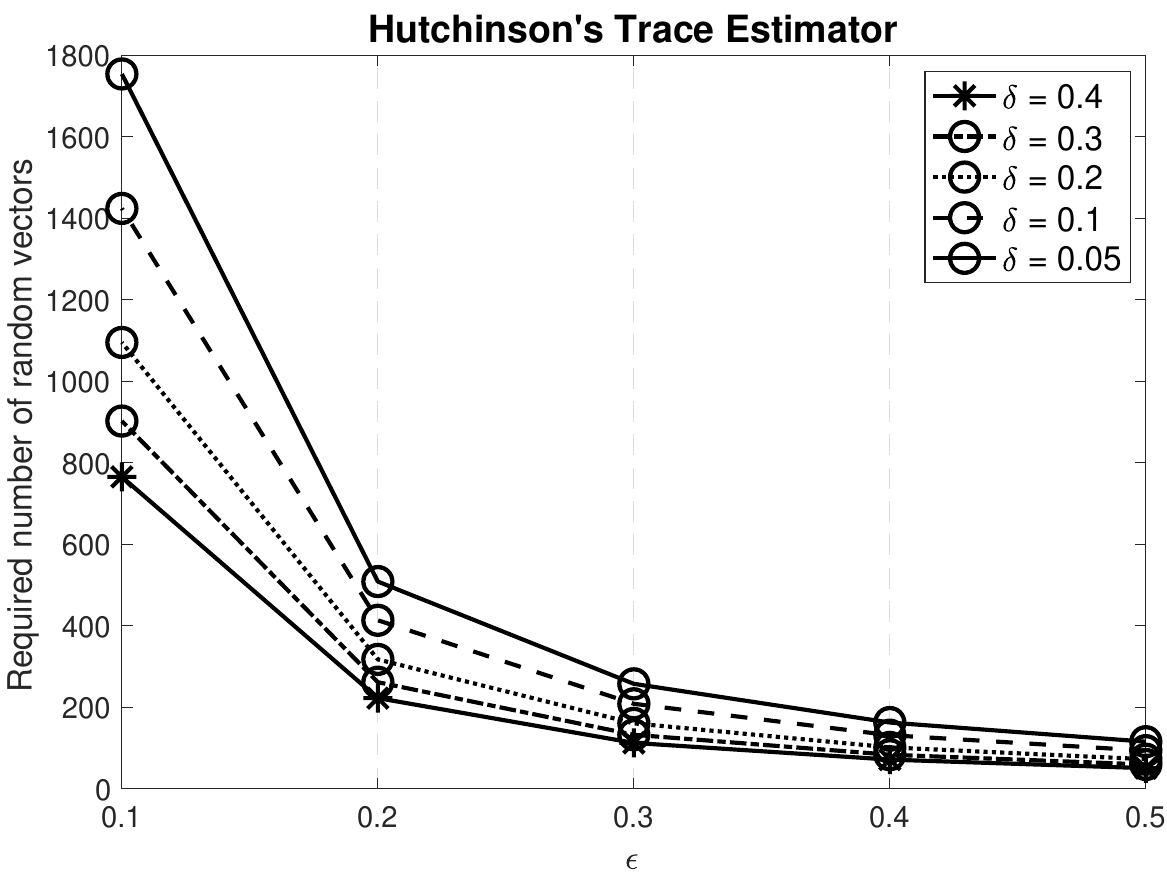}
  \caption{Required random vector counts based on different $\epsilon$ and $\delta$ values where $\epsilon$ is the worst case relative error which is guaranteed with a probability of 1-$\delta$}
  \label{fig:rvc} 
\end{figure}

Optimizing the number of random vectors would be crucial for the algorithm's efficiency.  For that, a statistical formula  $n_v(\epsilon,\delta) = \frac{2(2+\frac{8\sqrt{2}}{3}\epsilon)\text{log}(\frac{2}{\delta})}{\epsilon^2}$ proposed in~\citep{bib15}, gives the number of  random vectors that guarantees less than $\epsilon$ relative error with $1-\delta$ probability where $0< \delta < 1$. Fig.(\ref{fig:rvc}) represents the change of the number of random vectors with respect to the parameters $\epsilon$ and $\delta$.

\begin{algorithm}
\caption{Covariance Trace Estimator} \label{alg:tra}
\hspace*{\algorithmicindent} \textbf{Inputs:} $X_C$, 
$\epsilon$, $\delta$ \\
\hspace*{\algorithmicindent} \textbf{Output:}  
$\tau$
\begin{algorithmic}[1]
\State  $n_v \gets   \frac{2(2+\frac{8\sqrt{2}}{3}\epsilon)\text{log}(\frac{2}{\delta})}{\epsilon^2}$
\State \textit{sum} = 0
\For{\texttt{k=1:$n_v$}}
\State $z_k \text{ is constructed randomly using Eq. (\ref{eqn:trace})}$
\State $h_k = X_Cz_k $
\State $sum = sum + h_k^Th_k$

\EndFor
\State $\tau=sum/(n_v(N-1))$
\end{algorithmic}
\end{algorithm}

\subsection{Detection of Eigenvalue Search Intervals}
Although the trace estimation technique explained in the previous section can be used as an efficient tool for the estimation of the total variance of the data, the distribution of the eigenvalues of the covariance matrix should be also estimated efficiently. 
We employed a variant of Conjugate Gradient~(CG) algorithm. Although the original CG algorithm is proposed by Hestenes and Steifel in 1952 ~\citep{bib16}, there are many recent studies in literature to improve its scalability and robustness for exascale computing platforms ~\citep{yetkin01,yetkin02,greenbaum}.

The Ritz values obtained from the CG algorithm are identical to the eigenvalues of the input matrix in exact arithmetic due to the similarity of the matrices~\citep{bib17}. Therefore, we utilize the Ritz values that are obtained from Conjugate Gradient Least Square~(CGLS) algorithm, which is an extension of CG algorithm for over-determined linear systems and solves $X_C^TX_Cx=X_C^Tb$ by performing the matrix-vector  multiplication with the coefficient matrix, $X_C^TX_C$, implicitly. Hence, the Ritz values obtained from CGLS algorithm are same as the eigenvalues of $X_C^TX_C$ in exact arithmetic after $D$ iterations.

Since we require the eigenvalues of the $\textit{covariance matrix }, \mathcal{C}=\frac{X_C^TX_C}{N-1}$, we should run CGLS algorithm with $X_C$ matrix, and divide the estimated eigenvalues by $N-1$. Here, the right hand side vector $b$ is set to  \comment{$b=X_{C}^TX_{C}{\hat{x}}$ where ${\hat{x}}_j=1/\sqrt{D}$ such that $\|{\hat{x}}\|_2=1$} $b=X_C\textbf{1}/\|X_C\textbf{1}\|_2$ and $x_0=\textbf{0}$  where $\textbf{1}$ and $\textbf{0}$ are vectors of all  ones and zeros, respectively. Note that if the target problem is arising from a regression problem, the output vector can be used as the right hand side vector and the approximate solution can also be used as an initial guess. Also notice that we only require the $\alpha$ and $\beta$ values from the Alg.(\ref{alg:CGLS}); therefore, we can skip the computations of $x_i$s.

\begin{algorithm}
 \caption{Conjugate Gradient Least Squares} \label{alg:CGLS}
\hspace*{\algorithmicindent} \textbf{Input:} $X_C$, $b$, $x_0$, $maxiter$ \\
\hspace*{\algorithmicindent} \textbf{Output:}  $x^*$
\begin{algorithmic}[1]
\State $r_0=b-X_Cx_0$
\State $s_0=X_C^Tr_0$
\State $p_0=s_0$
\State $\gamma_0 = \|s_0\|_2^2$

\For{\texttt{i=0:$maxiter$}}
\State $q_i=X_Cp_i$
\State $\alpha_{i}=\gamma_i/\|q_i\|_2^2$
\State $x_{i+1}=x_{i}+\alpha_{i}p_{i}$
\State $r_{i+1}=r_{i}-\alpha_{i}q_i$
\State $s_{i+1}=X_C^Tr_{i+1}$
\State $\gamma_{i+1}=\|s_{i+1}\|_2^2$
\State $\beta_i=\gamma_{i+1}/\gamma_{i}$
\State $p_{i+1}=s_{i+1}+\beta_ip_{i}$
\EndFor
\State $x^* = x_{maxiter}$
\end{algorithmic}
\end{algorithm}

\begin{figure}
\centering
\includegraphics[width=0.9\textwidth]{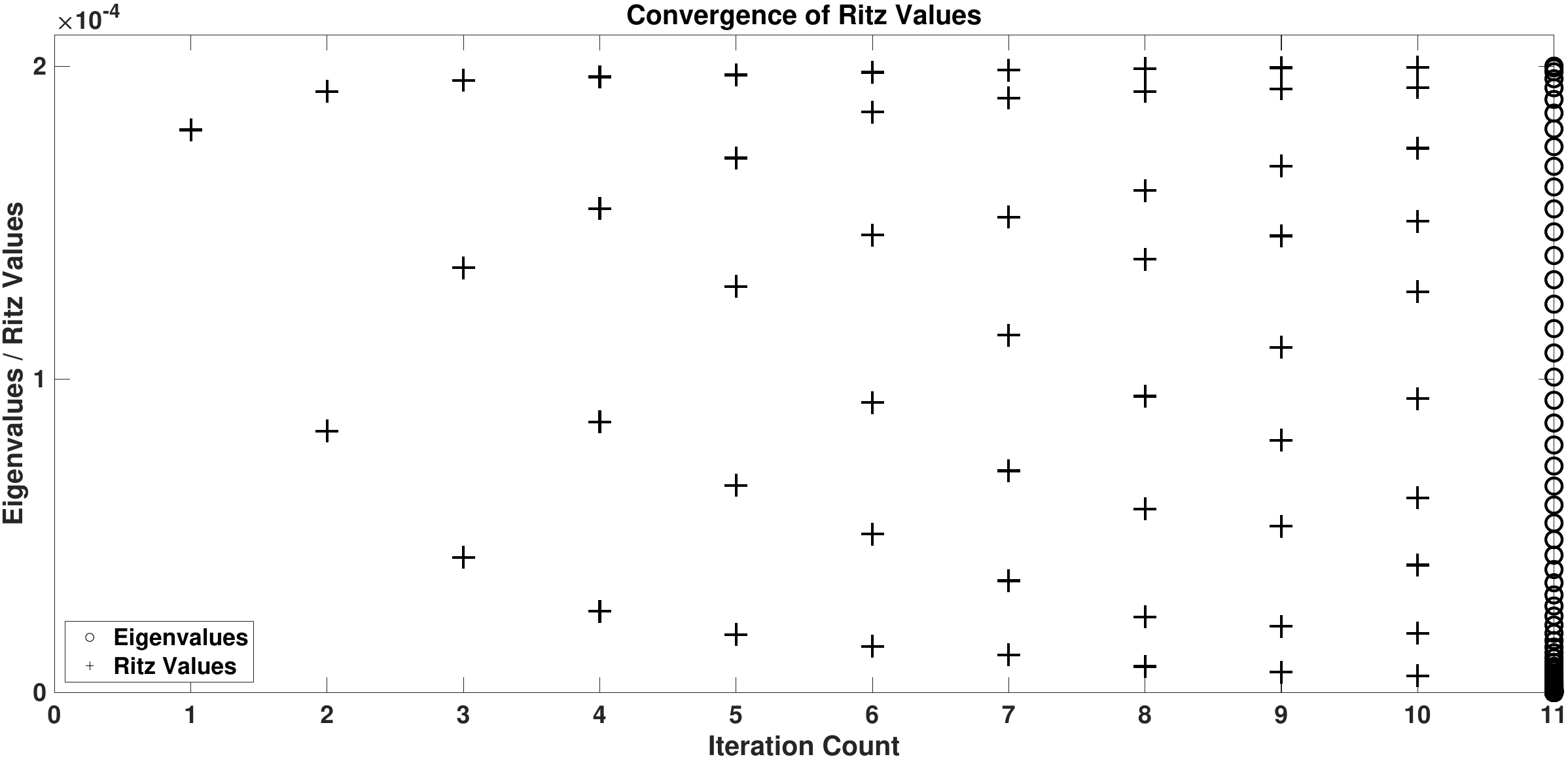}
\caption{The CGLS Algorithm is applied to the toy problem given in Section~\ref{mat}. 
The Eigenvalues / Ritz Values obtained after each iteration are given, and on the rightmost edge of the figure the exact eigenvalues of the covariance matrix are shown} 
\label{fig:1}

\end{figure}

Our goal is neither to find all eigenvalues of the $\textit{covariance matrix}$ nor to solve the least squares problem; instead, we want to approximate a few Ritz values to construct our eigenvalue search intervals; therefore, we iterate the CGLS only a few times and obtain a few required Ritz values. At this point, it should be noted that in a few iterations, the Ritz values firstly converge to the extreme, maximal and minimal, eigenvalues~\citep{bib20}. We confirm and illustrate this by obtaining the Ritz values at first 10 iterations from the CGLS algorithm applied to the toy problem given in Section~\ref{mat} as illustrated in Fig.(\ref{fig:1}). 

\comment{
In that step, we utilize the Ritz values that are obtained from Conjugate Gradient Least Square(CGLS) algorithm, which is an extension of Conjugate Gradient(CG) algorithm for over-determined linear systems. Let $\mathcal{B} \in R^{mxm}$ be \textit{symmetric, positive definite} matrix and $x,b \in R^m$. Then CG algorithm can be used to solve $\mathcal{B}x=b$ linear system\citep{bib16}.
\begin{algorithm}
 \caption{Conjugate Gradient} \label{alg:CG}
\hspace*{\algorithmicindent} \textbf{Input:} $\mathcal{B}$ = Data Matrix, $b$ = Target Vector, $x_0$ = Starting Vector \\
\hspace*{\algorithmicindent}  $maxiter$ = Maximum Iteration Number \\
\hspace*{\algorithmicindent} \textbf{Output:}  $x^*$ = Solution of Equation
\begin{algorithmic}[1]
\State $r_0=b-\mathcal{B}x_0$
\State $p_0=r_0$
\begin{equation*}
    T_{CG}=
    \begin{bmatrix} 
    \frac{1}{\alpha_0} & \frac{\sqrt{\beta_0}}{\alpha_0} & \cdots & \cdots & 0\\
    \frac{\sqrt{\beta_0}}{\alpha_0} & \frac{1}{\alpha_1}+\frac{\beta_0}{\alpha_0} & \frac{\sqrt{\beta_1}}{\alpha_1} & \cdots & 0\\
    \vdots & \frac{\sqrt{\beta_1}}{\alpha_1} & \frac{1}{\alpha_2}+\frac{\beta_1}{\alpha_1} & \ddots & 0\\
    \vdots & \vdots & \ddots & \ddots & \frac{\sqrt{\beta_{m-2}}}{\alpha_{m-2}}\\
   0 & 0 & 0 & \frac{\sqrt{\beta_{m-2}}}{\alpha_{m-2}} & \frac{1}{\alpha_{m-1}}+\frac{\beta_{m-2}}{\alpha_{m-2}}\\
    \end{bmatrix} 
\end{equation*}
\For{\texttt{i=0:\textit{maxiter}}}
\State $\alpha_{i}=\frac{r_{i}^Tr_{i}}{p_{i}^T\mathcal{B}p_{i}}$
\State $x_{i+1}=x_{i}+\alpha_{i}p_{i}$
\State $r_{i+1}=r_{i}-\alpha_{i}\mathcal{B}p_{i}$
\State $\beta_i=\frac{r_{i+1}^Tr_{i+1}}{r_{i}^Tr_{i}}$
\State $p_{i+1}=r_{i+1}+\beta_ip_{i}$
\EndFor
\State $x^* = x_{maxiter}$
\end{algorithmic}
\end{algorithm}
The eigenvalues of the tri-diagonal matrix $T_{CG}$, called Ritz values, are same as the eigenvalues of the matrix $\mathcal{B}$ in exact arithmetic after $m$ iterations because $T_{CG}$ and $\mathcal{B}$ are similar matrices\citep{bib17} .

However, the CG algorithm is conditioned to the fact that the $\mathcal{B}$ matrix is \textit{symmetric,positive definite}, therefore can not be used to solve a least square problem. To overcome that,assume we have rectangular data matrix $X$, an extension of CG called CGLS was proposed, which performs $X^T$ multiplication to both sides of the equation implicitly. Namely, CGLS solves $X^TXx=X^Tb$ \citep{bib16}. Also, to increase the algorithm's accuracy and minimize the effects of round-off errors, we use a re-orthogonalization version of CGLS. \\
\begin{algorithm}
 \caption{Conjugate Gradient Least Squares} \label{alg:CGLS}
\hspace*{\algorithmicindent} \textbf{Input:} $\mathcal{B}$ = Data Matrix, $b$ = Target Vector, $x_0$ = Initial Vector \\
\hspace*{\algorithmicindent}  $maxiter$ = Maximum Iteration Number \\
\hspace*{\algorithmicindent} \textbf{Output:}  $x^*$ = Solution of Equation
\begin{algorithmic}[1]
\State $r_0=b-\mathcal{B}x_0$
\State $s_0=\mathcal{B}^Tr_0$
\State $p_0=s_0$

\For{\texttt{i=0:\textit{maxiter}}}
\State $\alpha_{i}=\frac{s_{i}^T\mathcal{B}^T\mathcal{B}s_{i}}{p_{i}^T\mathcal{B}^T\mathcal{B}\mathcal{B}^T\mathcal{B}p_{i}}$
\State $x_{i+1}=x_{i}+\alpha_{i}p_{i}$
\State $r_{i+1}=r_{i}-\alpha_{i}\mathcal{B}p_{i}$
\State $s_{i+1}=\mathcal{B}^Tr_{i+1}$
\State $\beta_i=\frac{s_{i+1}^T\mathcal{B}^T\mathcal{B}s_{i+1}}{s_{i}^T\mathcal{B}^T\mathcal{B}s_{i}}$
\State $p_{i+1}=s_{i+1}+\beta_ip_{i}$
\EndFor
\State $x^* = x_{maxiter}$
\end{algorithmic}
\end{algorithm}
In that case, the eigenvalues of $T_{CGLS}$ matrix, Ritz values, are same as the eigenvalues of $X^TX$ in exact arithmetic after $D$ iterations. Since we require the eigenvalues of $\textit{covariance matrix } \mathcal{C}=\frac{X_C^TX_C}{N-1}$, we should run CGLS algorithm with $X_C$ matrix, and divide the estimated eigenvalues by $N-1$.

However, our goal is not to find all eigenvalues of the $\textit{covariance matrix}$; instead, we want to approximate a few ones to construct our eigenvalue search points; therefore, we iterate the CGLS a few times and construct a much smaller $T_{CGLS}$ matrix. At this point, it should be noted that in a few iterations, the Ritz values firstly converge to the extreme, maximal and minimal, eigenvalues\citep{bib20}.}

\subsection{Estimation of Eigenvalue Counts in an Intervals }
 Knowing the approximate number of the eigenvalues in a given interval can be very useful for load balancing, especially for large-scale eigenvalue calculations in parallel environments. It can also be beneficial to estimate the overall eigenvalue distribution of the covariance matrix. In this study, we will employ Chebyshev polynomial-based approaches for the approximate number of the eigenvalues in a specific interval determined by the few Ritz values of the covariance matrix to estimate the intrinsic dimension without constructing and solving the corresponding eigenproblem. The eigenvalue count $\eta_{[a,b]}$ of the matrix $\mathcal{A}$ in an interval [a,b] is equal to the trace of projector $\mathcal{P}$, which is constructed by the summation of outer products of the associated eigenvectors of the eigenvalues in the interval.
\begin{equation*}
    \eta_{[a,b]} = \text{tr}(\mathcal{P}) \text{ where } \mathcal{P} =\sum_{\lambda_i \in [a,b]} u_iu_i^T.
\end{equation*}
However, the projector can not be directly constructed because eigenvectors are unknown. Instead, it can be approximated by using the first kind Chebyshev polynomials if the input matrix is Hermitian as~\citep{bib18} \\

\begin{equation*}
    \mathcal{P} \approx \sum_{j=0}^p g_j^p\gamma_jT_i(\mathcal{A}).
\end{equation*}

Here, the well-known recursive relations between the first kind Chebyshev polynomials can be used to obtain the $p^{th}$ order Chebyshev polynomial such that $T_0(x) = 1$, $T_1(x) = x$ and  $T_{i+1}(x) = 2xT(x)_i - T(x)_{i-1}$. Note that, although the input is a matrix here, the recursion will be still valid.

The Jackson coefficient for smoothing the approximation is defined as
\begin{equation}
\label{eqn:Jackson}
    g_j^p = \frac{sin(j+1)\alpha_p}{(p+2)sin\alpha_p}+(1-\frac{j+1}{p+2})cos(j\alpha_p) \text{ where } \alpha_p=\frac{\pi}{p+2}.
\end{equation}
The coefficients of the Chebyshev polynomials are defined as
\begin{equation}
\label{eqn:gamma}
    \gamma_j =
    \begin{cases} 
      \frac{1}{\pi}(arccos(a)-arccos(b)) & j=0 \\
      \frac{2}{\pi}(\frac{sin(jarccos(a))-sin(jarccos(b))}{j}) & j > 0.\\
   \end{cases}
\end{equation}

The above approximation is based on the assumption that all eigenvalues of $\mathcal{A}$ lie in the [-1,1] interval. To extend it to a generic spectrum [$\lambda_{min}$,$\lambda_{max}$], the linear transformation in Eq.(\ref{eq:transform}) can be used, however $\lambda_{max}$ and $\lambda_{min}$ are not known. Therefore, their estimations can be used.  For the scheme to work accurately, the estimation for $\lambda_{max}$ and $\lambda_{min}$ must be greater than exact $\lambda_{max}$ and smaller than exact $\lambda_{min}$, respectively~\citep{bib18}.
\begin{equation}\label{eq:transform}
    l(t) = \frac{t-(\lambda_{max}+\lambda_{min})/2}{(\lambda_{max}-\lambda_{min})/2}
\end{equation}
Also, by applying the Hutchinson's Trace Estimator, the estimation can be done without constructing the approximated eigen-projector as~\citep{bib18}
\begin{equation*}
    \eta_{[a,b]} \approx
    \frac{1}{n_v}\sum_{k=1}^{n_v}\sum_{j=0}^pz_k^T g_j^p\gamma_jT_i(l(\mathcal{A}))z_k .
\end{equation*}
Also, to avoid the matrix-matrix multiplications in the definition of the Chebyshev polynomials, a new vector $w_j=T_j(l(\mathcal{A}))z$ can be defined. That transforms the 3-term recurrence of the Chebyshev Polynomials~\citep{bib18} to
\begin{equation*}
    w_{j+1} = 2l(\mathcal{A})w_j-w_{j-1}.
\end{equation*}

However, in the proposed method, we require the estimation for $\mathcal{C}=\frac{1}{N-1}X_{C}^TX_{C}$ matrix without constructing $\mathcal{C}$ explicitly since its computational and storage costs are expensive. Therefore, we should perform the action of matrix-matrix multiplications in the estimator implicitly to save computational time and storage. The linear transformation $l(t)$ should be applied to $\mathcal{C}$ which is also performed implicitly. 
If one consider the explicit form of the $\mathcal{C}$,  the above recursion can be re-written as 
\begin{equation}
\label{eqn:2}
    w_{j+1} = 2\biggl(\frac{2X_C^T(X_Cw_j)-\alpha (N-1)w_j}{\beta(N-1)}\biggr)-w_{j-1}
\end{equation}
where $\alpha = \lambda_{max} + \lambda_{min}$, $\beta = \lambda_{max} - \lambda_{min}$. Then, the number of eigenvalues of the covariance matrix can be estimated by
\begin{equation}
    \eta_{[a,b]} \approx
    \frac{1}{n_v}\sum_{k=1}^{n_v}\sum_{j=0}^pz_k^T g_j^p\gamma_jw_j.
\end{equation}

\begin{algorithm}
 \caption{Eigenvalue Count Estimator} \label{alg:eig}
\hspace*{\algorithmicindent} \textbf{Input:} $X_C$, $p$, $\epsilon$, $\delta$, $a$, $b$, $\lambda_{min}$, $\lambda_{max}$ \\
\hspace*{\algorithmicindent} \textbf{Output:}  $\eta_{[a,b]}$
\begin{algorithmic}[1]
\State $n_v \gets   \frac{2(2+\frac{8\sqrt{2}}{3}\epsilon)\text{log}(\frac{2}{\delta})}{\epsilon^2} $
\State $\alpha \gets \lambda_{max} + \lambda_{min}$
\State $\beta \gets \lambda_{max} - \lambda_{min}$
\State $a \gets \frac{2a-\alpha}{\beta}$
\State $b \gets \frac{2b-\alpha}{\beta}$
\State $sum=0$
\For{\texttt{k=1:\textit{$n_v$}}}
\State $z_k \text{ is constructed using Eq. (\ref{eqn:trace})}$
\For{\texttt{j=0:$p$}}
\State $g_j^p \text{ is constructed using Eq. (\ref{eqn:Jackson})} $
\State $\gamma_j \text{ is constructed using Eq. (\ref{eqn:gamma})} $
\State $w_j \text{ is constructed using Eq. (\ref{eqn:2})}$
\State $sum=sum+z_k^Tg_j^p\gamma_jw_j$

\EndFor
\EndFor
\State $\eta_{[a,b]}=sum/n_v$
\end{algorithmic}
\end{algorithm}
Hence, the eigenvalue count can be approximated by Alg.(\ref{alg:eig}), 
via two consecutive matrix-vector products, $X_Cw_j$ and $X_C^T(X_Cw_j)$ without constructing the covariance matrix explicitly.

The convergence rate of the proposed method is based on the gap between $a$,~$b$, the eigenvalue distribution of the input matrix because if either a or b is near or included in an eigenvalue cluster the estimation may be overestimated or underestimated, and the estimations for  $\lambda_{max}$ and  $\lambda_{min}$, and $p$ is recommended to be at least $70$~\citep{bib18}. 
However, in our numerical experiments, a smaller $p$ ($20$) produced acceptable results. 
It is because of the convergence of the Ritz values, and eigenvalue distribution of the covariance matrix of the data matrix. The biggest Ritz value fastly converges to the $\lambda_{max}$, which leads to better convergence for smaller p values, and gaps between middle Ritz values are bigger. Also, in the data covariance matrix, the biggest eigenvalues are much greater than the smallest eigenvalues, and generally smallest eigenvalues are clustered whereas the biggest are not and the proposed method does not care about the smallest eigenvalues.
\subsection{Proposed Method}

\begin{algorithm}[h]
 \caption{Proposed Method} \label{alg:PM}
\hspace*{\algorithmicindent} \textbf{Input:} $X_C$, $p$, $n_k$ ,$\epsilon$, $\delta$, $t_v$, $a_t$, $f_t$ \\
\hspace*{\algorithmicindent} \textbf{Output:}  $d$ 
\begin{algorithmic}[1]
\State $\tau \text{ is obtained using Alg. (\ref{alg:tra})}$
\State $(\mu_{n_k},\mu_1) \text{ is constructed using Alg. (\ref{alg:CGLS})}$
\State $n_v \gets   \frac{2(2+\frac{8\sqrt{2}}{3}\epsilon)\text{log}(\frac{2}{\delta})}{\epsilon^2}$
\State $\alpha \gets 0$
\State $d \gets 0$
\For{\texttt{i=1:$n_k$}}
\State $\eta_{[\mu_{i+1},\mu_{i}]} \text{ is obtained using Alg. (\ref{alg:eig})}$
\comment{
\If{$ a\_type == "mean"$}
\State $\alpha \gets \alpha + \frac{1}{2}*\eta_i*(\mu_{i+1}+\mu_{i})$
\Else
    \If {$ atype == "lower"$}
        \State $\alpha \gets \alpha + \eta_i*\mu_{i+1}$
    \Else {$ atype == "upper"$}
        \State $\alpha \gets \alpha + \eta_i*\mu_{i}$
    \EndIf
\EndIf
}
\State $\alpha_{[\mu_{i+1},\mu_{i}]} \text{ is computed based on } a_t $
\State $\alpha \gets \alpha + \alpha_{[\mu_{i+1},\mu_{i}]}$
\State $d \gets d + \eta_{[\mu_{i+1},\mu_{i}]}$
\If{$\alpha/\tau>t_v$}
        \State break;
\EndIf
\EndFor
\comment{
\If {$ rtype == "ratio"$}
    \State $\textit{Ratio the last interval for extra variance}$
\Else {$ rtype == "iter"$}
    \State $\textit{Continue the algorithm by shortening the last interval to find more accurate result}$
\EndIf
}
\State $d \text{ is finalized based on } f_t $
\end{algorithmic}
\end{algorithm}

In the proposed method, first we estimate the trace of the covariance matrix $\tau$ via Alg.(\ref{alg:tra}) and get a few Ritz values $(\mu_{n_k},\mu_1)$ where $\mu_1$ is the largest via Alg.(\ref{alg:CGLS}). Then, starting with the largest Ritz value interval, we estimate the number of eigenvalues $\eta_{[\mu_{2},\mu_{1}]}$ in that interval via Alg.(\ref{alg:eig}) (on line 7). At this point, we have the  lower, upper bound, and estimated count of the eigenvalues. Then, the question is how one can approximate the summation of eigenvalues in the interval $\alpha_{[\mu_{2},\mu_{1}]}$ . For that, we have three options indicated by the $a_t$ parameter in the pseudocode (on line 8) which can either multiply the estimated eigenvalue count with the lower bound, the upper bound, or their mean. 
The numerical results have shown that the mean approximation for the eigenvalue summation is the most accurate one. Later, we check if the variance for that interval is bigger than the target variance $t_v$ (on line 11), if not we proceed with the smaller interval and check for the cumulative variance. These iterations continue until we obtain a bigger approximated variance than $t_v$. Once we obtain a bigger approximated variance, we have three options to conclude the procedure, determined by $t_f$ (on line 15)\comment{( {\color{red} [diğer yerlerde de algorithmaların satır numaralarıne refer edelim]}}. To finalize $d$, we can return the estimated ID; we can assume that eigenvalues are distributed linearly in the latest interval, and we can estimate the ID based on that assumption; or we can shorten our intervals and continue our method with them.  

\subsection{Practical Considerations }
For the approximations of $\lambda_{min}$ and $\lambda_{max}$, we use 0 and $c_1\mu_{1}$ respectively, where $c_1$ is a randomly chosen constant greater than 1. 0 is used as an approximation for the smallest eigenvalue because it is known that eigenvalues of the covariance matrix are larger than 0. Here, it should be remembered that the Alg.(\ref{alg:eig})'s convergence with respect to $p$ is dependent on the approximations for $\lambda_{min}$ and $\lambda_{max}$. Therefore, interpreting the smallest Ritz Value and finding an approximation for the smallest eigenvalue of the covariance matrix and better approximation for $\lambda_{max}$ can improve the convergence performance of the proposed method.

At the first iteration of the algorithm, we run the Alg.(\ref{alg:eig}) for the $\eta_{[\mu_{1},c_2\mu_{1}]}$ interval where $c_2$ is a randomly chosen constant greater than 1 to approximate the number of eigenvalues of the covariance matrix which are larger than the largest Ritz value. 

It is important that if the result of Alg.(\ref{alg:eig}) (on line 7) of proposed method is 1, it indicates that only the lower bound exists in that interval; therefore, lower approximation should be used. 

Since generally finding exactly $80\%$ variance is not necessary, we defined a parameter acceptable range $a_r$. We do not perform finalization step if $t_v+a_r>\alpha/\tau>t_v-a_r$ (on line 11). 

The proposed method suggested acceptable results 
for linear data; however, to improve its performance for nonlinear data, we performed several clustering approaches to the data such as well-known k-means and spectral clustering algorithms. Then, ID estimated for each cluster by applying the proposed approach 
and took the arithmetic mean of each cluster's estimated ID. Clustering algorithms create an additional computational cost, especially if spectral clustering \citep{ng2001spectral, pasadakis2022multiway} is used. However,  they can be parallelized \citep{manguoglu2010tracemin}  and also one can easily recycle the computed neighborhood matrices for the non-linear data in the manifold learning application.

\subsection{Cost Analysis}
The proposed method only depends on the matrix-vector products; therefore, it has $O(NDn_kn_vp)$ time complexity. Moreover, since $n_k$, $np$, and $p$ are weakly dependent on $N$ and $D$ and are also relatively small, the time complexity is $O(ND)$. Table (\ref{tab:comparison}) compares the time complexities of different ID estimators. The proposed method has the lowest time complexity.

\begin{table}[h]
    \centering
    \begin{tabular}{l|l}
         \textbf{Method}&\textbf{Time Complexity}  \\
         \hline
         Proposed Method & $O(ND)$\\
         MLE & $O(N^2D)$ \\
         $\text{MiND}_{\text{MLi}}$ & $O(N^2D^2)$ \\
         $\text{MiND}_{\text{MLk}}$ & $O(N^2D^2)$ \\
         $\text{MiND}_{\text{KL}}$ & $O(N^2D^2)$ \\
         DANCo & $O(N\text{log}(N)D^2)$ \\
         FastDANCo & $O(N\text{log}(N)D)$ \\
         ESS & $O(N^{d+1}D(d+1)^2)$
    \end{tabular}
    \caption{Time complexities of different ID estimators}
    \label{tab:comparison}
\end{table}

Moreover, the random vector multiplications for trace estimation in Alg.(\ref{alg:tra}) and Alg.(\ref{alg:eig}) are independent of each other hence they can be easily parallelized to speed up the running time of the proposed method. Also, in  Alg.(\ref{alg:eig}) the random vector multiplications can be computed incrementally, and after each increment the change of the result can be checked to stop the algorithm~\citep{bib18}. Therefore, there is a high probability that the algorithm can terminate in small number of iterations.

\section{Numerical Experiments}\label{sec4}

To demonstrate the advantages and limits of the proposed approach, we used both artificially generated TF-IDF data and well-known synthetic benchmark data from the ID estimation literature. Numerical experiments are performed on an Intel Core i7 PC with 16GB memory using Matlab.

\subsection{Synthetic Data} \label{mat}
For the below experiments, a $5.000\times500$ matrix is created by the Scikit-Learn library make-low-rank function which creates a low-rank matrix with bell-shaped singular values~\citep{scikit-learn}; in real life, this type of matrix can be seen in gray-level pictures of faces or TF-IDF vectors of text documents crawled from the web~\citep{scikit-learn}. In our experiments, effective rank and tail strength of the matrix are $30$ and $0,05$, respectively. In the matrix, dimensions 20, 21, 22 and 23 corresponds to $78.16\%$,  $80.29\%$, $82.24\%$ and $84.03\%$ variances respectively.   

\begin{figure}[h!]
  \includegraphics[width=\textwidth]{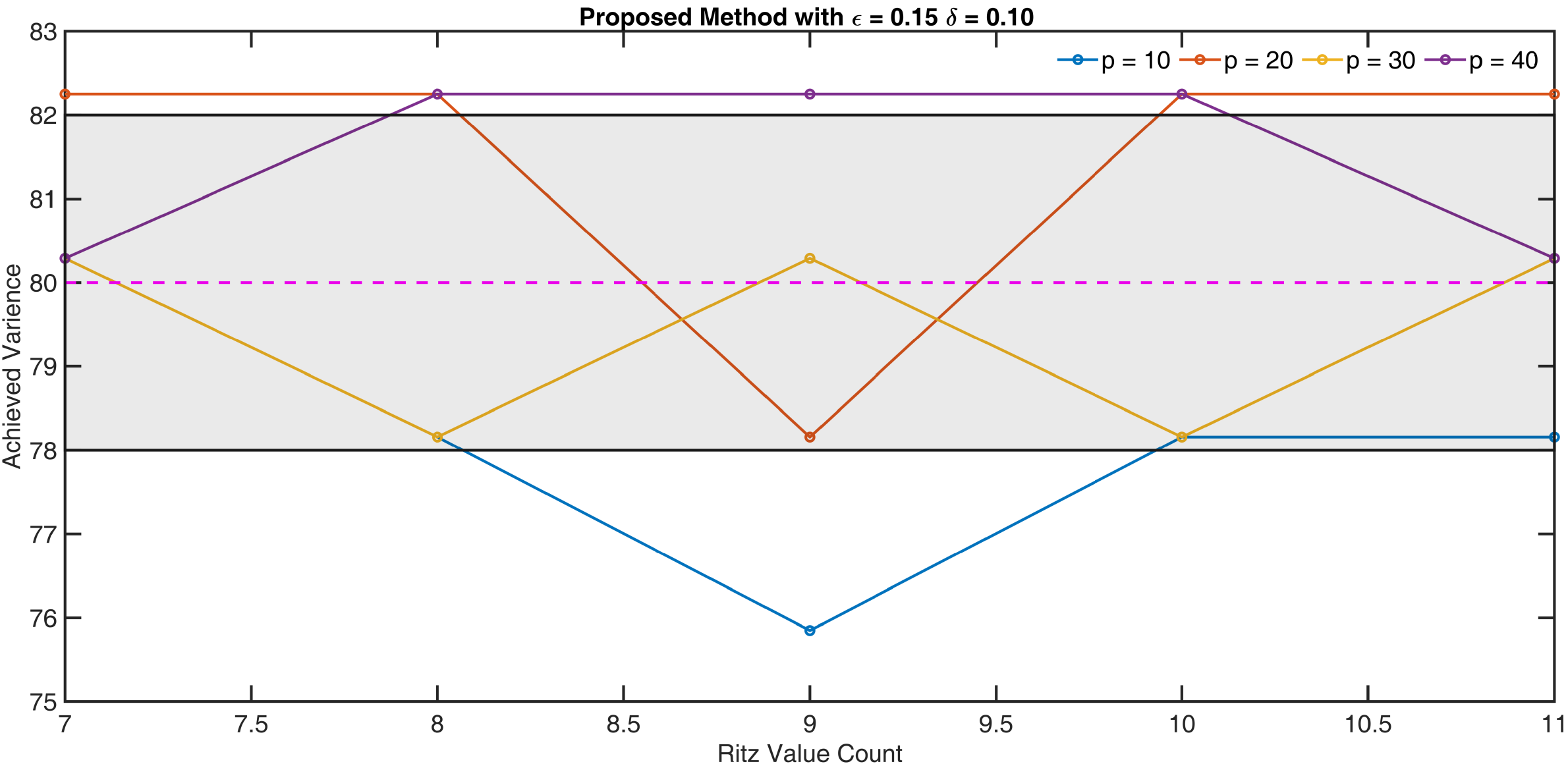}
  \caption{Effect of the number of the Chebyshev polynomials (p) and the Ritz value counts on the total variance estimation where $\epsilon$ and $\delta$ are fixed  (consequently the $n_v$ is also fixed). An acceptable range of the variance is given as a shaded area in the figure}
  \label{fig:PR} 
\end{figure}

We target $80\%$ variances in our test cases since it is a reasonable selection for the total variance in most of the cases. However, it is an adaptable parameter for the proposed approach. We defined the $a_r$ as $2\%$, $c_1$ as $1.5$, $c_2$ as $1.4$, $a_t$ as mean, and $f_t$ as linear. As shown in Fig.(\ref{fig:PR}), number of the Chebyshev polynomials should be at least 20 for reasonable variance estimation. On the other hand, number of the Ritz intervals only causes some predictable fluctuations on the results.

\begin{figure}[h!]
  \includegraphics[width=\textwidth]{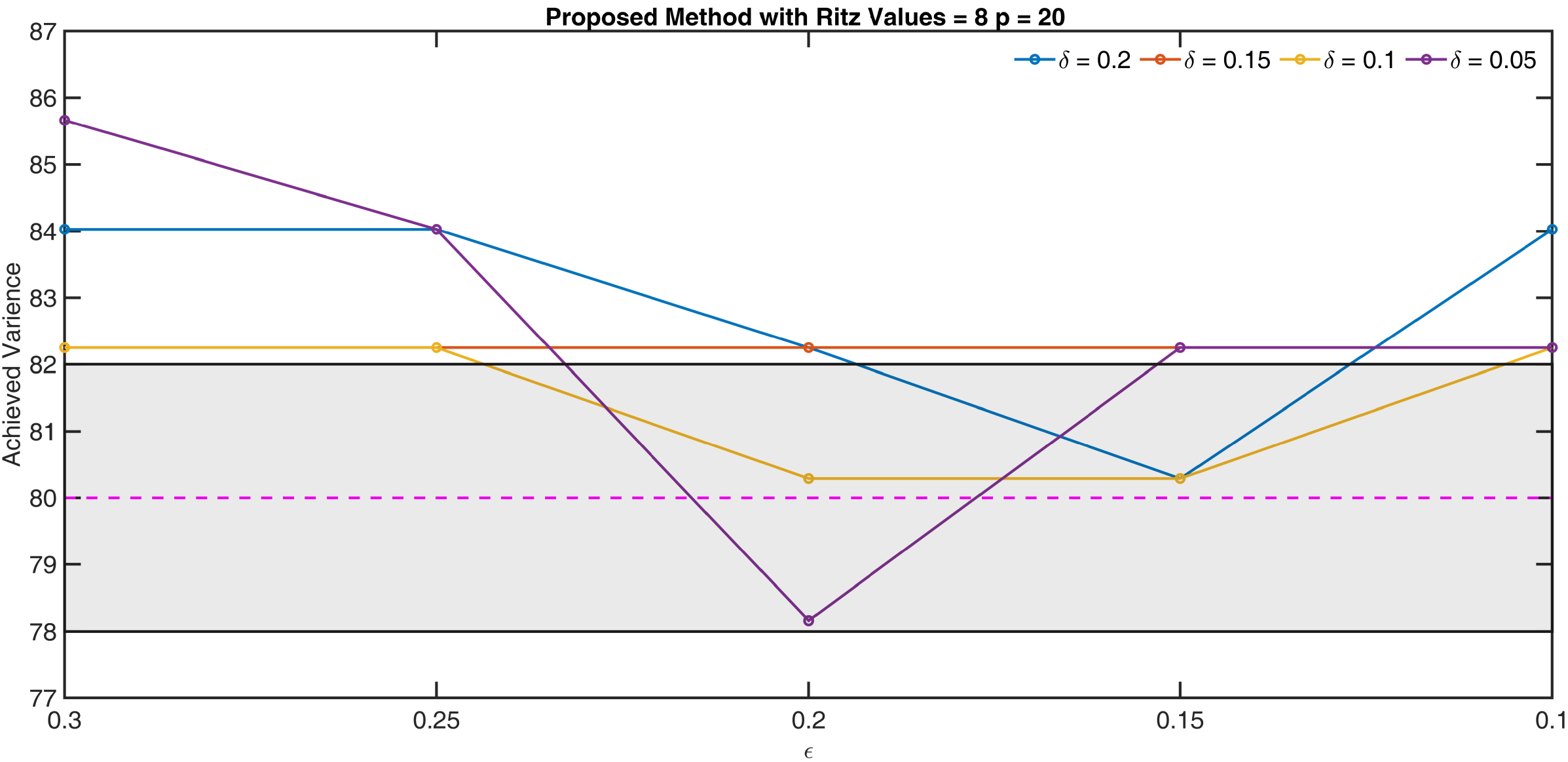}
  \caption{
  Effect of the $\epsilon$ and $\delta$ (consequently the $n_v$) on the total variance estimation where p and Ritz values are fixed. An acceptable range of the variance is given as a shaded area in the figure}
  \label{fig:ED} 
\end{figure}

To measure the effect of the $\epsilon$ and $\delta$ parameters, we have used fixed values for the $p$ and number of the Ritz intervals as 20 and 8 respectively by using the results shown in Fig.(\ref{fig:ED}).    

As expected, decreasing $\epsilon$ parameter for Hutchinson approximation makes the ID estimation more stable and more accurate. Although this improvement increase the overall computational time of the algorithm, this extra cost can easily be handled in high-performance computing systems due to the embarrassingly parallel nature of the Alg.(\ref{alg:tra}) and Alg.(\ref{alg:eig}). Note that, all calculations in Alg.(\ref{alg:PM}) is only based on matrix-vector multiplications and can be easily realized in parallel environments.

\begin{figure}[h!]
  \includegraphics[width=\textwidth]{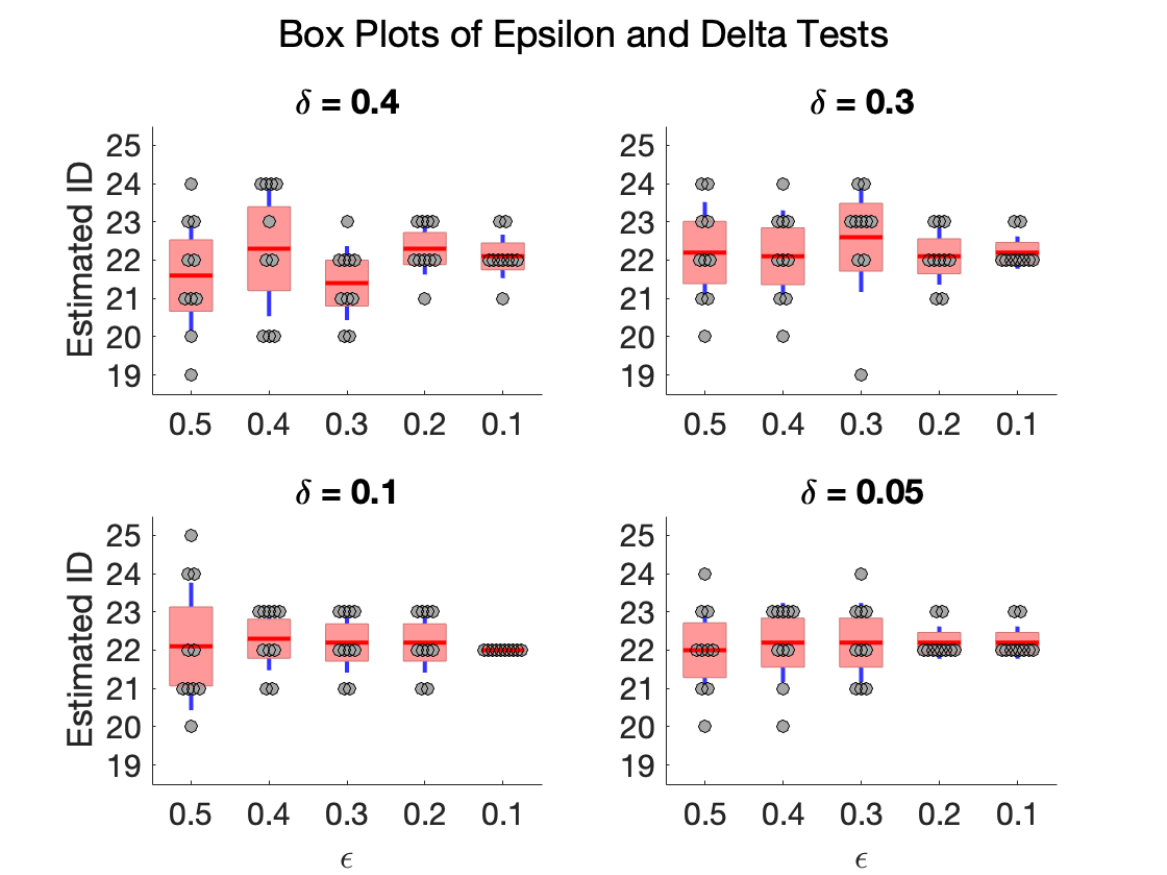}
  \caption{Estimated ID has been obtained for every $\epsilon$-$\delta$ pair 10 times and their statistical distributions are presented where gray disk, red line, light red box, and blue line correspond to the result of an individual run, the median, the interquartile range, and non-outlier range of the corresponding runs, respectively    
  }
  \label{fig:boxPlotPoints} 
\end{figure}

Since the proposed method contains randomness, we tested each epsilon and delta configuration 10 times and 
Fig.(\ref{fig:boxPlotPoints}) shows the statistical distribution of the individual tests via box-plots. Note that, in Fig.(\ref{fig:boxPlotPoints}), instead of showing the estimated variance, we depicted the estimated ID. It can be concluded that as the $n_v$ increase, the results get more stable with the cost of increasing the number of random vectors using the trace and projector estimation. 

\comment{
\subsubsection{Iterating over Epsilon and Delta Values}
\includegraphics[width=\textwidth]{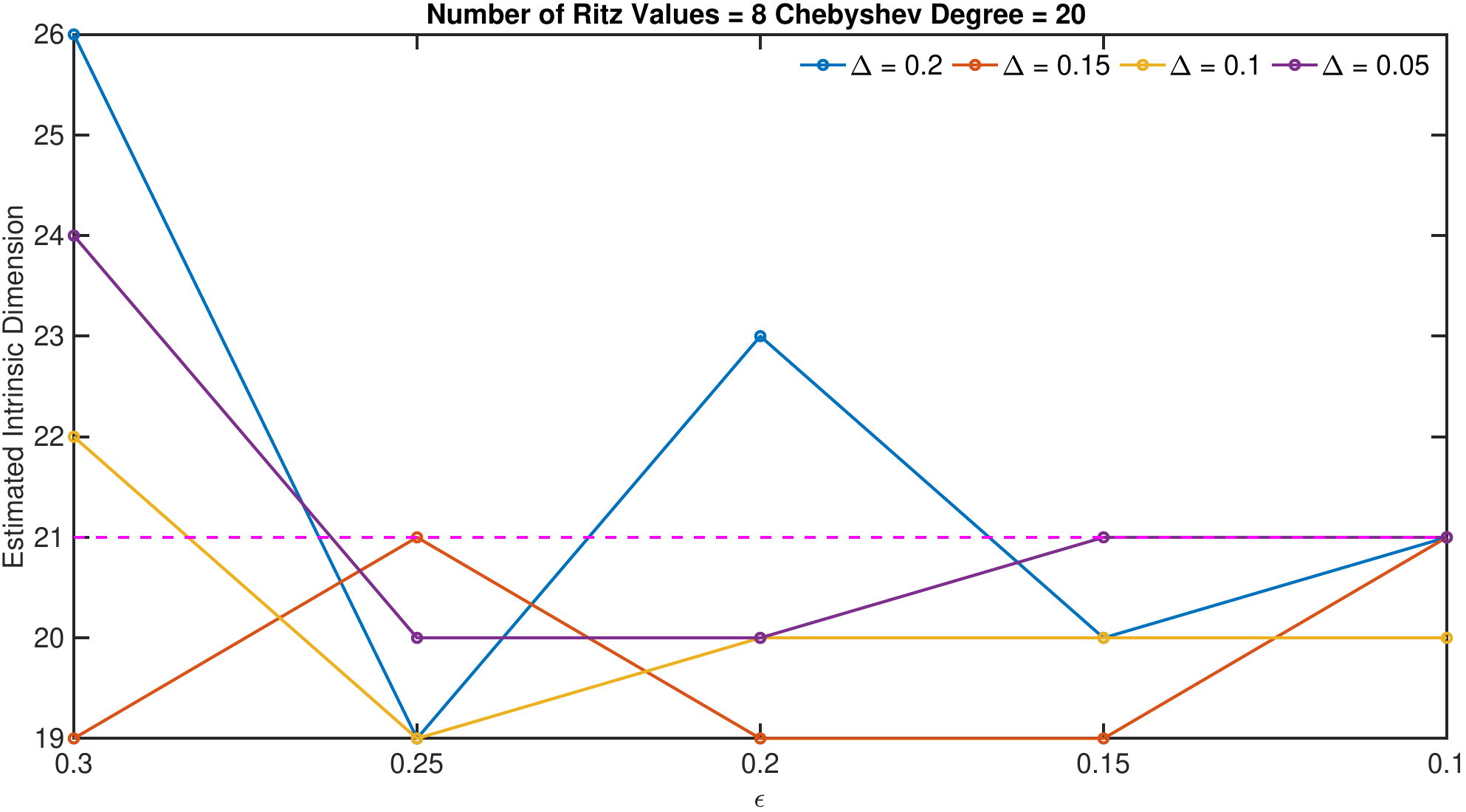}
\includegraphics[width=\textwidth]{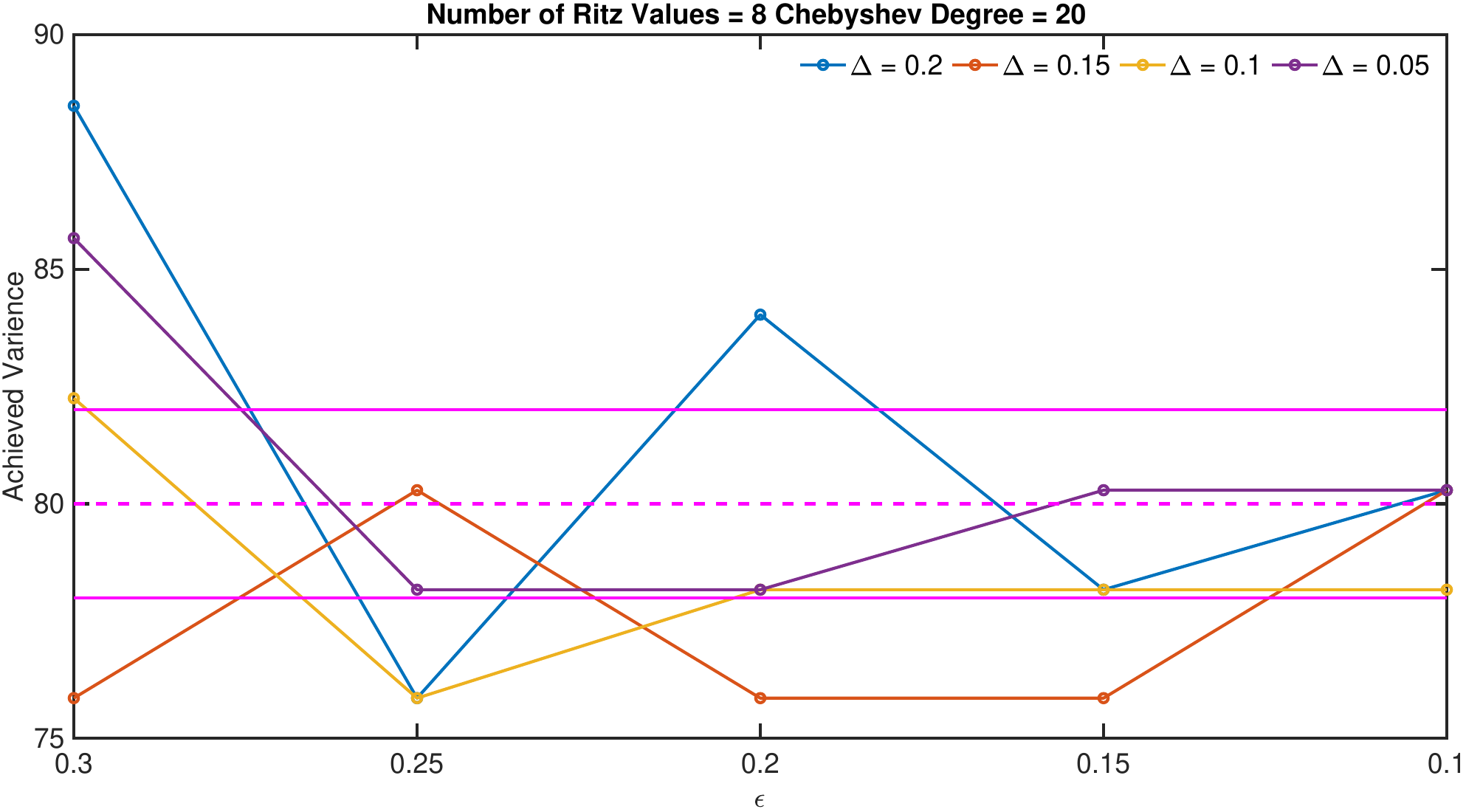}
\subsubsection{Iterating over Ritz Count and Chebyshev Polynomial Degree}
\includegraphics[width=\textwidth]{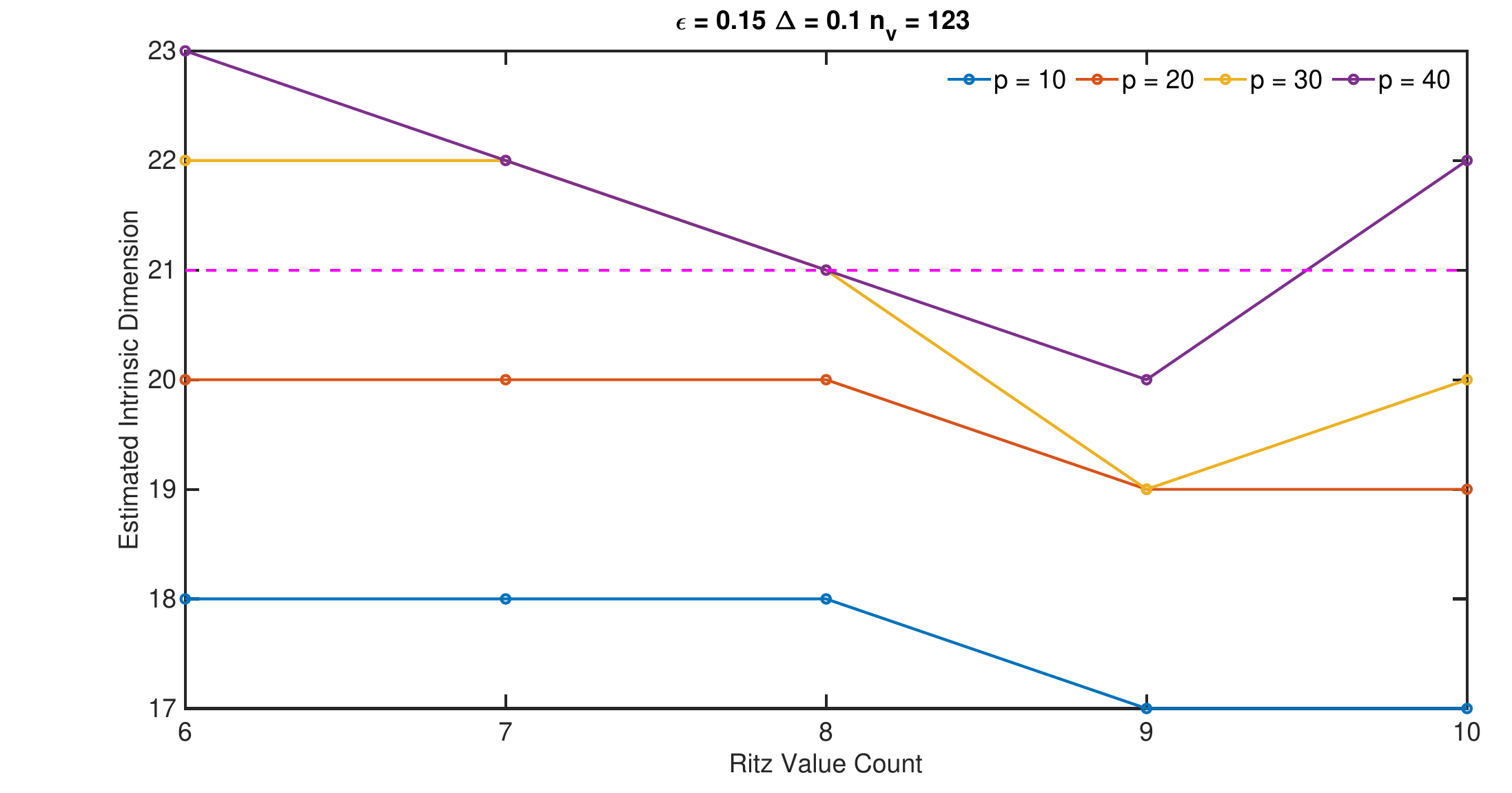}
\includegraphics[width=\textwidth]{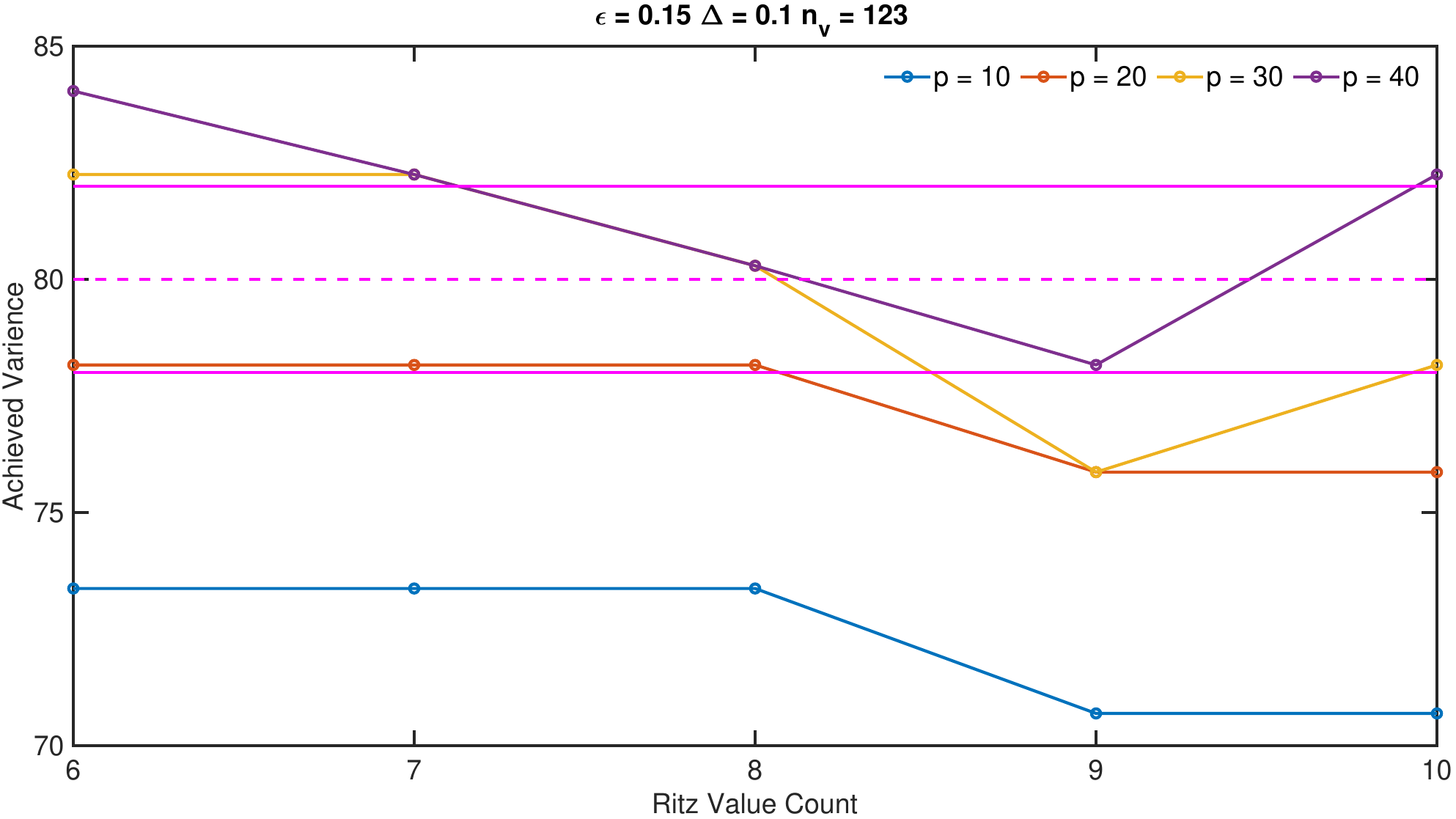}
}

\subsection{Benchmark Data}
For the generation of the synthetic data sets we tested, the Scikit-Dimension library BenchmarkManifolds function~\citep{scikit-dimension}, which creates 24 different data set based on~\citep{datasets}, was used. The parameters of the proposed method were 20, 8, 0.2, and 0.2 for $p$, $n_k$, $\epsilon$, and $\delta$ respectively and cluster size for variants of the proposed method was 2. We also tested a wide variety of intrinsic dimension estimators to show the efficiency of our method. All these results are given in Table {\ref{tab:allResults}}. The main observation from these results, the proposed method is giving competitive results concerning most approaches we have tested. The last row of the Table {\ref{tab:allResults}} gives the mean of the relative estimation errors, which is defined as the difference between the ID value of the dataset and the estimated one divided correspond. If these results are examined carefully, one can conclude that the proposed approach with k-means clustering has a lower estimation error than the other projection-based ID estimators.
On the other hand, topological estimators mostly have better estimations. Nevertheless, as discussed in section 2, topological estimators predominantly suffer from high computational costs. Therefore, especially for the large datasets, the proposed approach can be used as a fast and reliable estimator. Note that all estimators were tested by using the functions in~\citep{scikit-dimension} with their default values, and no parameter optimizations were conducted. Notice that for $\mathcal{M}_{10d}$ dataset, the result of FSH method is "nan". As the authors pointed out, that is because of the implementation in~\citep{scikit-dimension}. Hence, we excluded $\mathcal{M}_{10d}$ from the FSH method tests. 

\begin{figure}[h!]
  \centering
\begin{minipage}[b]{.47\textwidth}
  \includegraphics[width=.95\linewidth]{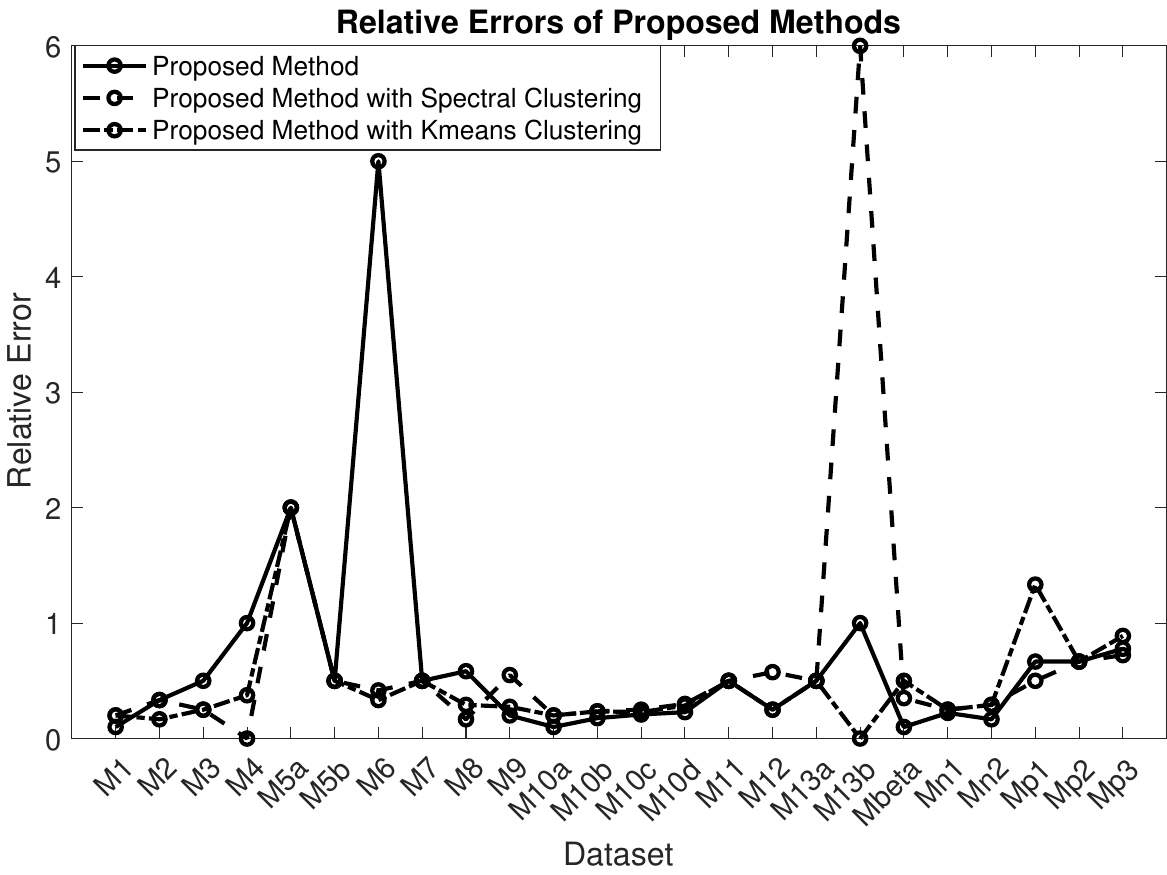}  \\
   a) All datasets 
  \label{fig:6.2.a}
\end{minipage}\hfill
\begin{minipage}[b]{.47\textwidth}
  \includegraphics[width=.95\linewidth]{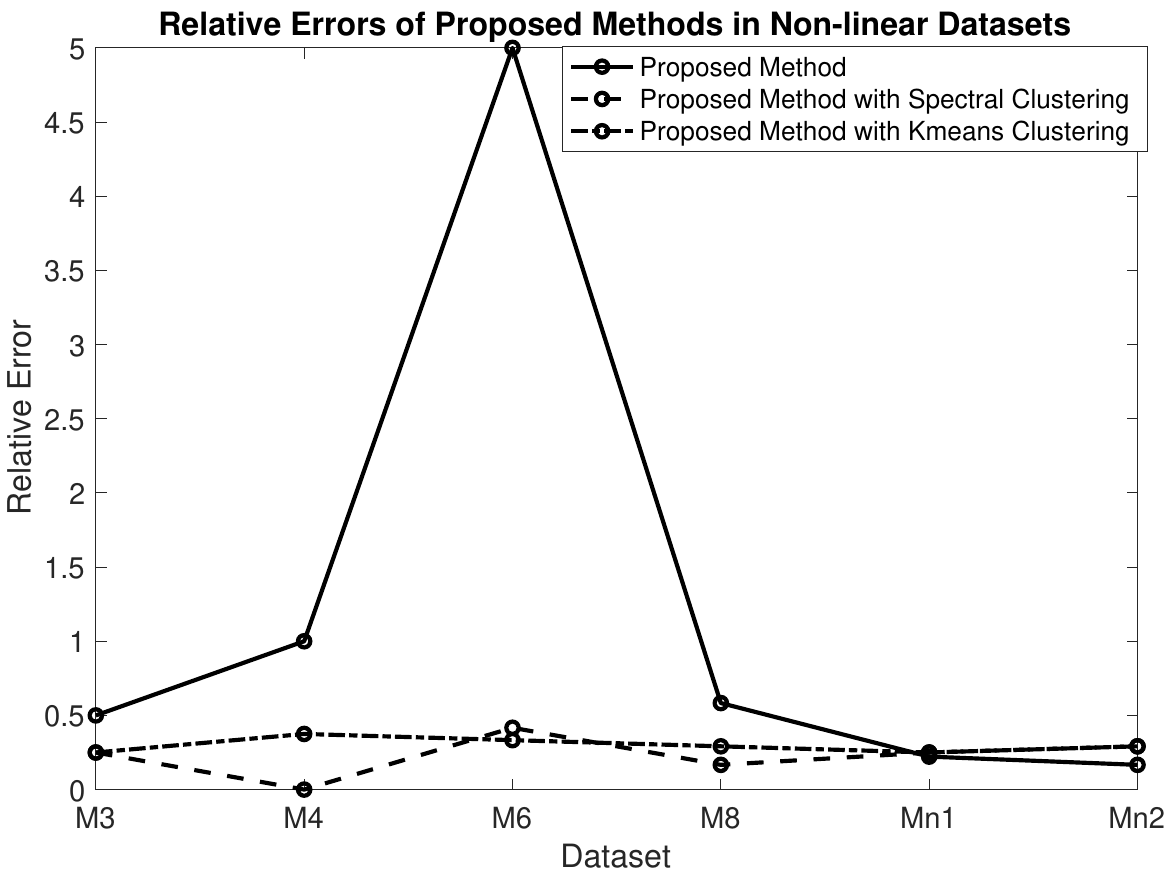}  \\
  b) Non-linear Datasets 
  \label{fig:6.2.b}
\end{minipage}\hfill
\caption{Relative errors of the variants of the proposed methods for all and only non-linear datasets}
\label{fig::6}
\end{figure}

We have focused on the proposed approach in the second representation of the results. In Fig.(\ref{fig::6}.a), the proposed method and its variants are compared based on the relative dimension estimation errors. Although the general trend of all variants is having an error of less than one for most of the problems in the datasets with all three variants in $\mathcal{M}_{5a}$ the error is large for all of them. This dataset contains three features and represents a 1-D Helix structure. Its highly non-linear structure and the low number of dimensions is the leading cause of this situation. Another two crucial peak points of the figures are $\mathcal{M}_{6}$ for the vanilla version of the proposed method and $\mathcal{M}_{13b}$ for the proposed method with spectral clustering. Again this worst case occurs due to the nature of the benchmark data. $\mathcal{M}_{6}$ is a highly non-linear manifold data; therefore, the vanilla version cannot capture the non-linear nature of the data.
On the other hand, the expected intrinsic dimension of $\mathcal{M}_{13b}$ is only one, while its number of features is 13 with only 3 of them greater than 0. Therefore, Ritz values unpredictably converged, they suggested some unreal eigenvalues between 0 and the lowest eigenvalue of $\mathcal{M}_{13b}^T\mathcal{M}_{13b}$ which causes the greater suggestion of ID. It should be noted that this is related to the randomness of the proposed method because different tests on this dataset yielded different results, for instance, while Ritz value count 3, the $PM_{spec}$ suggested the ID as 1.5 and 2. However, these are not reflected in the table to preserve the wholeness of the testing parameters. Moreover, the removal of $\mathcal{M}_{13b}$ from the tests of $PM_{spec}$ nearly reduces its mean relative error to 0.44 which is very competitive with other methods. Also, it can be seen that none of the proposed methods can even approximately capture the ID of the $\mathcal{M}_p$ family which is non-linearly embedded paraboloids. In Fig.(\ref{fig::6}.b), the clear advantage of clustering approaches for the highly non-linear datasets is depicted. Hence, it can be concluded that the estimation accuracy of the proposed method and variants are depended on the shape of a dataset. 

\section{Conclusion}
Large-scale data analysis usually needs dimensionality reduction application to reduce computational time and improve the accuracy of the target machine learning applications. However, without an apriori knowledge of intrinsic dimension, it can be a costly operation. This study proposes a novel and computationally efficient intrinsic dimension estimation approach based on Ritz values and Chebyshev polynomials. The proposed method is a projective, but its main advantage is its ability to estimate the intrinsic dimension without constructing and solving the underlying eigenvalue problem. We have performed several tests on various problems ranging from linear to highly non-linear structures. The performance of the proposed approach is compared with other well-known estimators in the literature. 

Furthermore, the dominant computational kernel of the method is based on only matrix-vector multiplication operations, and it is parallelizable inherently. Therefore it is a proper candidate for large-scale data analysis on parallel computing platforms.

\section{Declarations}
{\bf{Funding:}}
This research is funded as a part of "A Novel Approach for Scalable Manifold Learning"   (\href{http://sca-ml.khas.edu.tr/}{for more details}) project under the project number 120E281 within the framework 1001 by TUBITAK(The Scientific and Technological Research Council of Turkey). Other than that, the authors have no relevant financial, ethical, or non-financial interests to disclose.

{\bf{Author Contributions:}}
All authors contributed to the study conception, methodology, and design. Algorithmic design, data curation and visualization were performed by Kadir Özçoban, Murat Manguoğlu and E. Fatih Yetkin. The first draft of the manuscript was written by Kadir Özçoban and all authors commented on previous versions of the manuscript. All authors read and approved the final manuscript, and consent for publication.

{\bf{Data and Code Availability:}}
We obtained the data from a publicly available repository/generator, as described in Section 4.2. The code is developed as a part of an ongoing project but can be shared upon request.
\comment{
\begin{table}[!h]
    \centering
      \caption{Nomenclature}
    \begin{tabular}{|c|c|}
    \hline
        Symbol & Mean \\
    \hline
        $N$ & Sample Size \\
    \hline
    $D$ & Feature Size \\
    \hline
         $X \in R^{NxD}$ & Data Matrix \\
    \hline     
         $X_c$ & Centered Data Matrix \\
    \hline     
        $\mathcal{C}$ & Covariance \\
    \hline
        $n_v$ & random vector count\\
    \hline
    $z_k$ & random vectors\\
    
    \hline
        $\tau$ & Trace of $\mathcal{C}$ \\
    \hline
        $\eta_i$ & Eigenvalue count \\
    \hline
        $\mu_i$ & Ritz Values \\
    \hline
        $n_k$ & Number of Ritz Value \\
    \hline
        $\alpha$ & Cumulative Variance \\
    \hline
            $t_v \in [0,1]$ & Target Variance for Method \\
    \hline
    $a_t$ & Approximationg type: lower, mean, upper \\
    \hline
    $f_t$ & Finalization type \\
    \hline
    $iid$ & Independent and Identically Distrubited \\
    \hline
    $a_r$ & Acceptable Range \\
    \hline
    \end{tabular}
    \label{tab:my_label}
\end{table}
}

\newpage

\begin{sidewaysfigure}[ht]
\centering
\begin{adjustbox}{width=\textwidth}
    
\begin{tabular}{|c|c|c|c|c|c|c|c|c|c|c|c|c|c|c|c|c|c|c|c|c|c|c|}
\hline
Dataset & ID & PM & $PM_{Spec}$ & $PM_{Kmeans}$ &PFO & PFN & PMG & PRT & PPR & PKS & PBS & CID & FSH & KNN & MADA & $Mind_{ML}$ & MLE & MOM & TLE & TNN & DANCo & ESS \\ 
\hline
$\mathcal{M}_1$ & \textbf{10.00} & 9.00 & 8.00 & 8.00 & 11.00 & 7.00 & 1.00 & 1.00 & 10.94 & 5.00 & 1.00 & 8.91 & 11.02 & 11.00 & 9.62 & 9.10 & 9.53 & 8.60 & 10.10 & 9.28 & 11.00 & 10.18 \\ 
\hline
$\mathcal{M}_2$ & \textbf{3.00} & 2.00 & 2.00 & 3.50 & \textbf{3.00} & 2.00 & \textbf{3.00} & 1.00 & 2.27 & 2.00 & \textbf{3.00} & 2.87 & 2.92 & 4.00 & 3.07 & 2.90 & 3.03 & 2.79 & 3.13 & 2.92 & 2.96 & 2.90 \\ 
\hline
$\mathcal{M}_3$ & \textbf{4.00} & 2.00 & 3.00 & 5.00 & 5.00 & 2.00 & 2.00 & 1.00 & 2.75 & 2.00 & 3.00 & 3.58 & 3.34 & 1.00 & 4.14 & 3.77 & 4.03 & 3.77 & 4.21 & 3.79 & 4.67 & 4.02 \\ 
\hline
$\mathcal{M}_4$ & \textbf{4.00} & 8.00 & \textbf{4.00} & 5.50 & 8.00 & 5.00 & 1.00 & 1.00 & 7.96 & \textbf{4.00} & 1.00 & 3.75 & 5.81 & \textbf{4.00} & 4.34 & 3.86 & 4.18 & 4.49 & 4.42 & 3.74 & 5.04 & 5.03 \\ 
\hline
$\mathcal{M}_{5a}$ & \textbf{1.00} & 3.00 & 3.00 & 3.00 & 3.00 & 2.00 & 2.00 & \textbf{1.00} & 2.40 & 2.00 & \textbf{1.00} & 1.01 & 2.91 & 3.00 & 1.08 & \textbf{1.00} & 1.06 & 1.17 & 1.11 & 0.98 & \textbf{1.00} & 1.63 \\ 
\hline
$\mathcal{M}_{5b}$ & \textbf{2.00} & 3.00 & 3.00 & 3.00 & 3.00 & \textbf{2.00} & \textbf{2.00} & 1.00 & 2.23 & \textbf{2.00} & 1.00 & 2.79 & 2.91 & 1.00 & 3.17 & 2.37 & 2.82 & 2.86 & 2.91 & 1.97 & 2.47 & 2.88 \\ 
\hline
$\mathcal{M}_6$ & \textbf{6.00} & 36.00 & 8.50 & 8.00 & 12.00 & 9.00 & 12.00 & 1.00 & 11.95 & 12.00 & 1.00 & \textbf{6.00} & 8.43 & 2.00 & 7.34 & 6.25 & 6.91 & 7.16 & 7.12 & 5.87 & 8.76 & 8.52 \\ 
\hline
$\mathcal{M}_7$ & \textbf{2.00} & 3.00 & 3.00 & 3.00 & 3.00 & \textbf{2.00} & 1.00 & 1.00 & 2.95 & 1.00 & 1.00 & 1.96 & 2.88 & 3.00 & 2.09 & \textbf{2.00} & 2.07 & 2.01 & 2.14 & 1.94 & 2.21 & 2.15 \\ 
\hline
$\mathcal{M}_8$ & \textbf{12.00} & 19.00 & 10.00 & 15.50 & 24.00 & 17.00 & 24.00 & 2.00 & 23.77 & 24.00 & 1.00 & 11.33 & 17.95 & 5.00 & 14.43 & 10.00 & 14.24 & 13.21 & 13.83 & 13.80 & 11.99 & 19.50 \\ 
\hline
$\mathcal{M}_9$ & \textbf{20.00} & 16.00 & 9.00 & 14.50 & \textbf{20.00} & 11.00 & 19.00 & 1.00 & 19.86 & 9.00 & 1.00 & 13.48 & 19.07 & 3.00 & 15.12 & 10.00 & 15.14 & 13.49 & 15.25 & 14.77 & 19.00 & 19.41 \\ 
\hline
$\mathcal{M}_{10a}$ & \textbf{10.00} & 11.00 & 8.00 & 8.00 & 11.00 & 6.00 & 2.00 & 1.00 & 10.97 & 5.00 & 1.00 & 8.58 & 10.43 & 11.00 & 9.15 & 8.89 & 9.20 & 8.39 & 9.62 & 9.10 & 11.00 & 10.30 \\ 
\hline
$\mathcal{M}_{10b}$ & \textbf{17.00} & 14.00 & 13.00 & 13.00 & 18.00 & 10.00 & \textbf{17.00} & 1.00 & 17.88 & 8.00 & 1.00 & 12.80 & 17.10 & 4.00 & 13.96 & 10.00 & 13.90 & 12.47 & 14.20 & 13.96 & 18.00 & 17.29 \\ 
\hline
$\mathcal{M}_{10c}$ & \textbf{24.00} & 19.00 & 18.00 & 18.50 & 25.00 & 14.00 & 23.00 & 2.00 & 24.76 & 14.00 & 1.00 & 16.41 & 23.27 & 5.00 & 17.87 & 10.00 & 17.97 & 15.90 & 18.01 & 18.05 & 24.01 & 24.38 \\ 
\hline
$\mathcal{M}_{10d}$ & \textbf{70.00} & 54.00 & 49.00 & 50.00 & 71.00 & 39.00 & 69.00 & 3.00 & 69.05 & 35.00 & 1.00 & 31.62& \textbf{nan} & 71.00 & 36.65 & 10.00 & 37.17 & 31.47 & 34.71 & 39.85 & 71.00 & 69.95 \\ 
\hline
$\mathcal{M}_{11}$ & \textbf{2.00} & 3.00 & 3.00 & 3.00 & 3.00 & 1.00 & \textbf{2.00} & 1.00 & 2.15 & \textbf{2.00} & 1.00 & 1.97 & 1.97 & \textbf{2.00} & 2.10 & \textbf{2.00} & 2.07 & 2.21 & 2.16 & 2.02 & 2.26 & 2.49 \\ 
\hline
$\mathcal{M}_{12}$ & \textbf{20.00} & 15.00 & 8.50 & 15.00 & \textbf{20.00} & 11.00 & 11.00 & 1.00 & 19.82 & 11.00 & 1.00 & 12.74 & 20.30 & 6.00 & 16.04 & 10.00 & 16.22 & 13.98 & 15.47 & 17.04 & 12.05 & 19.79 \\ 
\hline
$\mathcal{M}_{13a}$ & \textbf{2.00} & 3.00 & 3.00 & 3.00 & 3.00 & 1.00 & 1.00 & 1.00 & 1.86 & 1.00 & \textbf{2.00} & 1.94 & 2.88 & 1.00 & 2.08 & \textbf{2.00} & 2.06 & 1.93 & 2.13 & 1.96 & 2.23 & 2.05 \\ 
\hline
$\mathcal{M}_{13b}$ & \textbf{1.00} & 2.00 & 7.00 & \textbf{1.00} & 2.00 & \textbf{1.00} & 3.00 & \textbf{1.00} & 2.03 & 2.00 & 3.00 & 4.01 & 1.97 & \textbf{1.00} & 3.78 & 1.20 & 1.75 & 1.83 & 1.84 & \textbf{1.00} & \textbf{1.00} & 1.95 \\ 
\hline
$\mathcal{M}_{beta}$ & \textbf{10.00} & 9.00 & 6.50 & 5.00 & \textbf{10.00} & 7.00 & 20.00 & 1.00 & 10.62 & \textbf{10.00} & 11.00 & 3.47 & 5.27 & 40.00 & 6.64 & 6.31 & 6.63 & 5.78 & 6.21 & 6.54 & 8.02 & 5.92 \\ 
\hline
$\mathcal{M}n_1$ & \textbf{18.00} & 14.00 & 13.50 & 13.50 & 27.00 & 13.00 & 36.00 & 1.00 & 18.82 & \textbf{18.00} & 19.00 & 12.58 & 17.00 & 11.00 & 14.31 & 10.00 & 14.26 & 12.71 & 14.29 & 14.24 & 21.00 & 18.39 \\ 
\hline
$\mathcal{M}n_2$ & \textbf{24.00} & 20.00 & 17.00 & 17.00 & 36.00 & 18.00 & 48.00 & 1.00 & 25.08 & \textbf{24.00} & 25.00 & 15.42 & 23.30 & 16.00 & 17.87 & 10.00 & 17.85 & 15.63 & 17.49 & 18.26 & 31.00 & 24.85 \\ 
\hline
$\mathcal{M}p_1$ & \textbf{3.00} & 1.00 & 1.50 & 7.00 & 1.00 & 1.00 & 11.00 & 1.00 & 1.24 & 1.00 & 2.00 & 2.14 & 0.90 & 1.00 & 3.07 & 2.90 & 3.04 & 2.73 & 3.13 & 2.99 & 3.20 & 2.85 \\ 
\hline
$\mathcal{M}p_2$ & \textbf{6.00} & 2.00 & 2.00 & 2.00 & 1.00 & 21.00 & 20.00 & 1.00 & 1.06 & 1.00 & 2.00 & 2.63 & 0.89 & 21.00 & 5.14 & 5.08 & 5.19 & 4.22 & 5.34 & 5.48 & 7.01 & 4.89 \\ 
\hline
$\mathcal{M}p_3$ & \textbf{9.00} & 2.00 & 15.50 & 1.00 & 1.00 & 1.00 & 29.00 & 1.00 & 1.03 & 1.00 & 2.00 & 2.92 & 0.89 & 30.00 & 6.55 & 6.52 & 6.64 & 4.76 & 6.70 & 7.23 & 8.94 & 6.12 \\ 
\hline
 $Mean_{error}$ &  &  0.68 &  0.68 &  0.46 &  0.50 &  0.49 &  0.82 &  0.73 &  0.39 &  0.49 &  0.62 &  0.36 &  0.39 &  0.77 &  0.29 &  0.23 &  0.18 &  0.25 &  0.20 &  0.12 &  0.13 &  0.20 \\ 
\hline

\end{tabular}
\end{adjustbox}
\caption{The relative errors for all datasets are obtained using different ID estimators. The mean of the relative errors for each estimator is given in the last row. The ID of the datasets is given in the first column. If an estimator obtains the exact ID, then the corresponding value is emphasized in bold.}

\label{tab:allResults}
\end{sidewaysfigure}

\clearpage

\comment{
\section{Cost Analysis Notes}
For an input matrix $X^{NxD}$, the complexity of construction of covariance matrix is $O(ND^2)$, and the complexity of its eigen-decompositon is $O(D^3)$, hence the overall complexity of PCA is $O(ND^2+D^3)$.

In our proposed method ($\ref{alg:PM}$), for line 1 we $O(n_vND)$ complexity where $O(ND)$ and $O(N)$ comes from $5^{th}$ and $6^{th}$ line of Alg($\ref{alg:tra}$) respectively . 

For line 2 in ($\ref{alg:PM}$), we have $O(n_kND)$ complexity from CGLS ($\ref{alg:CGLS}$) where for line 6 we have $O(ND)$, for line 7 we have $O(N)$, for line 8 we have $O(D)$, for line 9 we have $O(N)$, for line 10 we have $O(ND)$, for line 11 we have $O(D)$ and for line 12 we have $O(D)$. Also, to find Ritz values we should solve eigen-decomposition of $n_kxn_k$ matrix which has $O(n_k^3)$ complexity.

For line 7 in ($\ref{alg:PM}$), we have $O(n_vpND)$ complexity from ($\ref{alg:eig}$) where for line 12 we have $O(ND+ND+D+N)$, for line 12 we have $O(N)$.

Therefore, the overall complexity of ($\ref{alg:PM}$) is $O(n_vND+n_kND+n_k^3+n_kn_vpND) \approx O(n_kn_vpND)$ where $n_k < 10, n_v < 150, p < 100$. However, these approach can just be used to determine intrinsic dimension. To adapt it to the full PCA after finding the $d$, one can use partial svd.

\includegraphics[width = \textwidth]{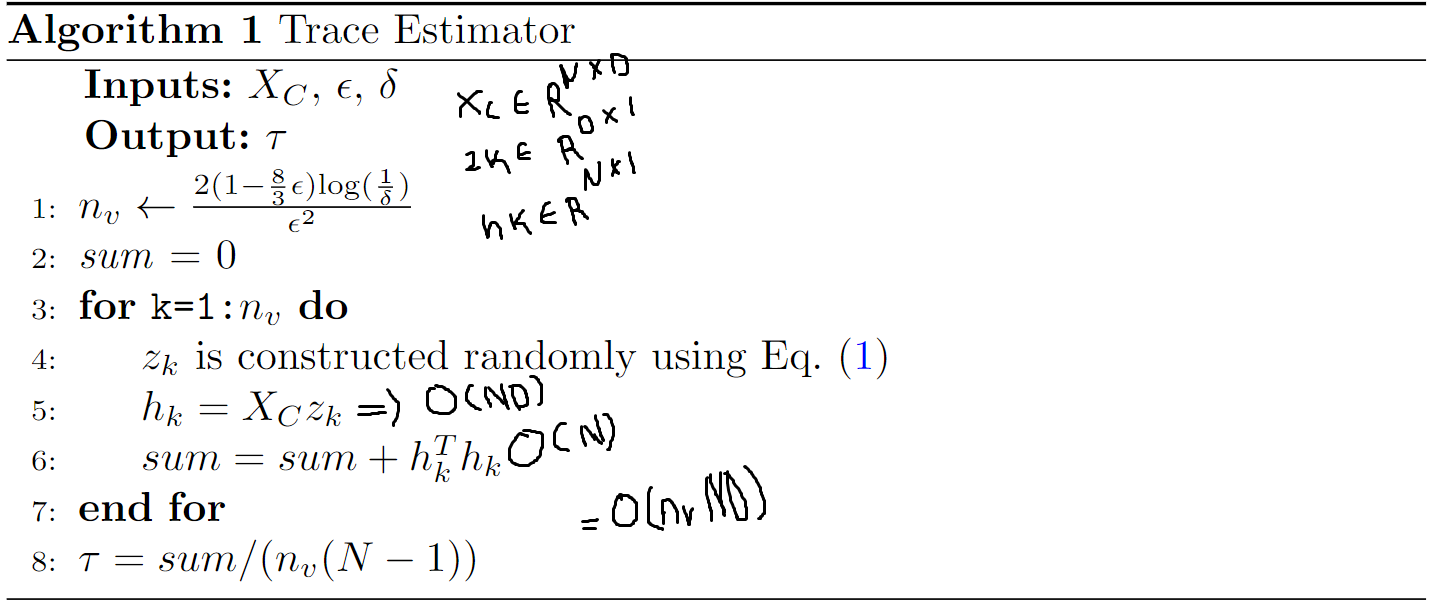}
\includegraphics[width = \textwidth]{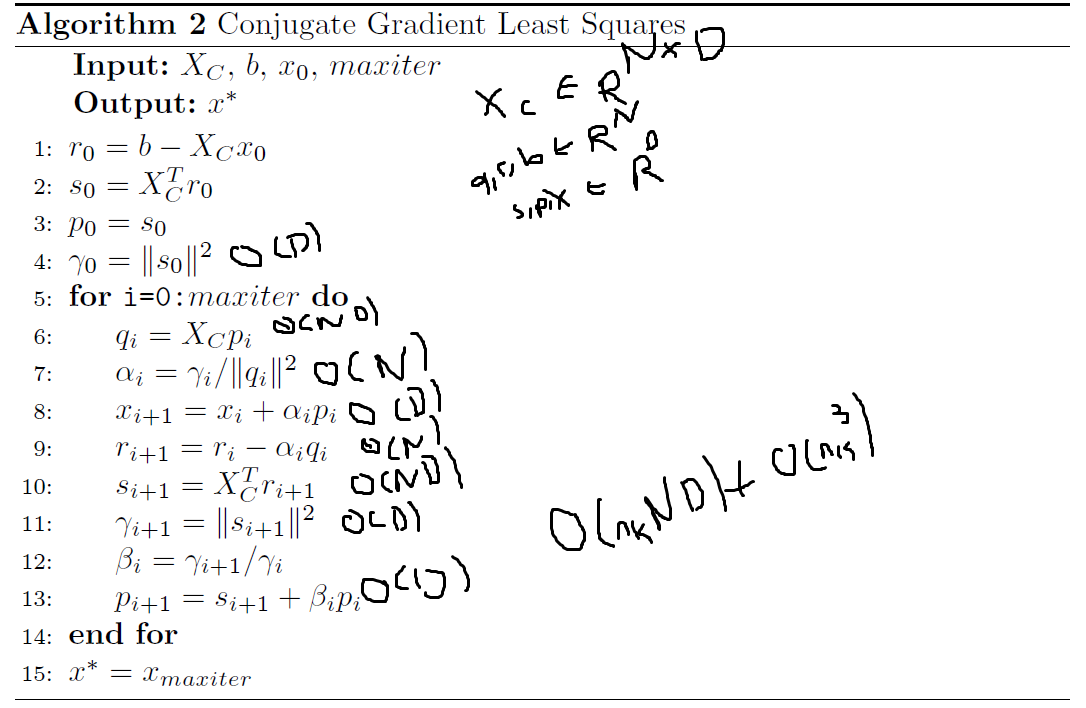}
\includegraphics[width = \textwidth]{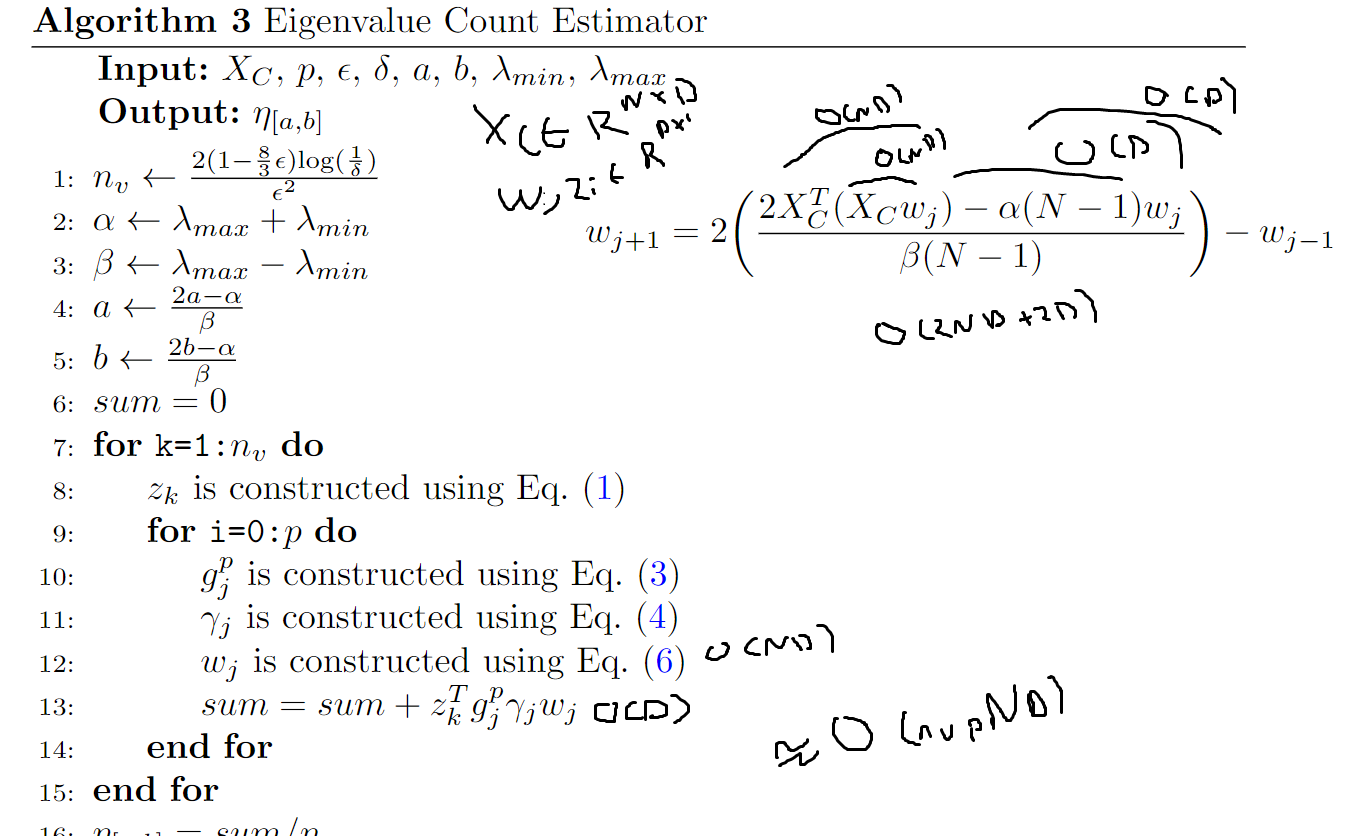}
\includegraphics[width = \textwidth]{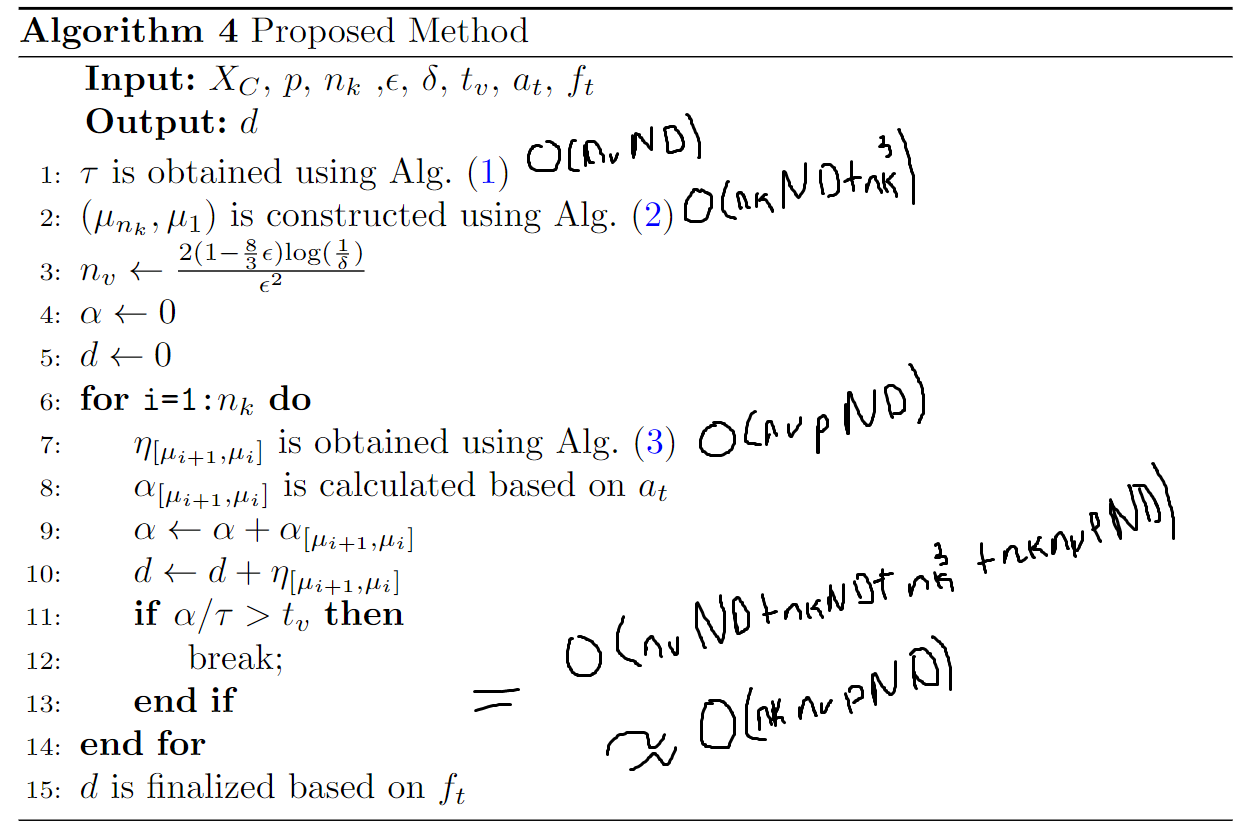}
\newpage}

\comment{

\subsubsection{MNIST Data}
MNIST is a handwritten digits dataset that contains $60.000$ trains and $10.000$ test samples. We obtain $10.000$ test samples from the dataset and estimate the intrinsic dimension with various parameters. Later, we perform the dimensionality reduction to our figure into the suggested dimension for each test using PCA.
\includegraphics[width=\textwidth]{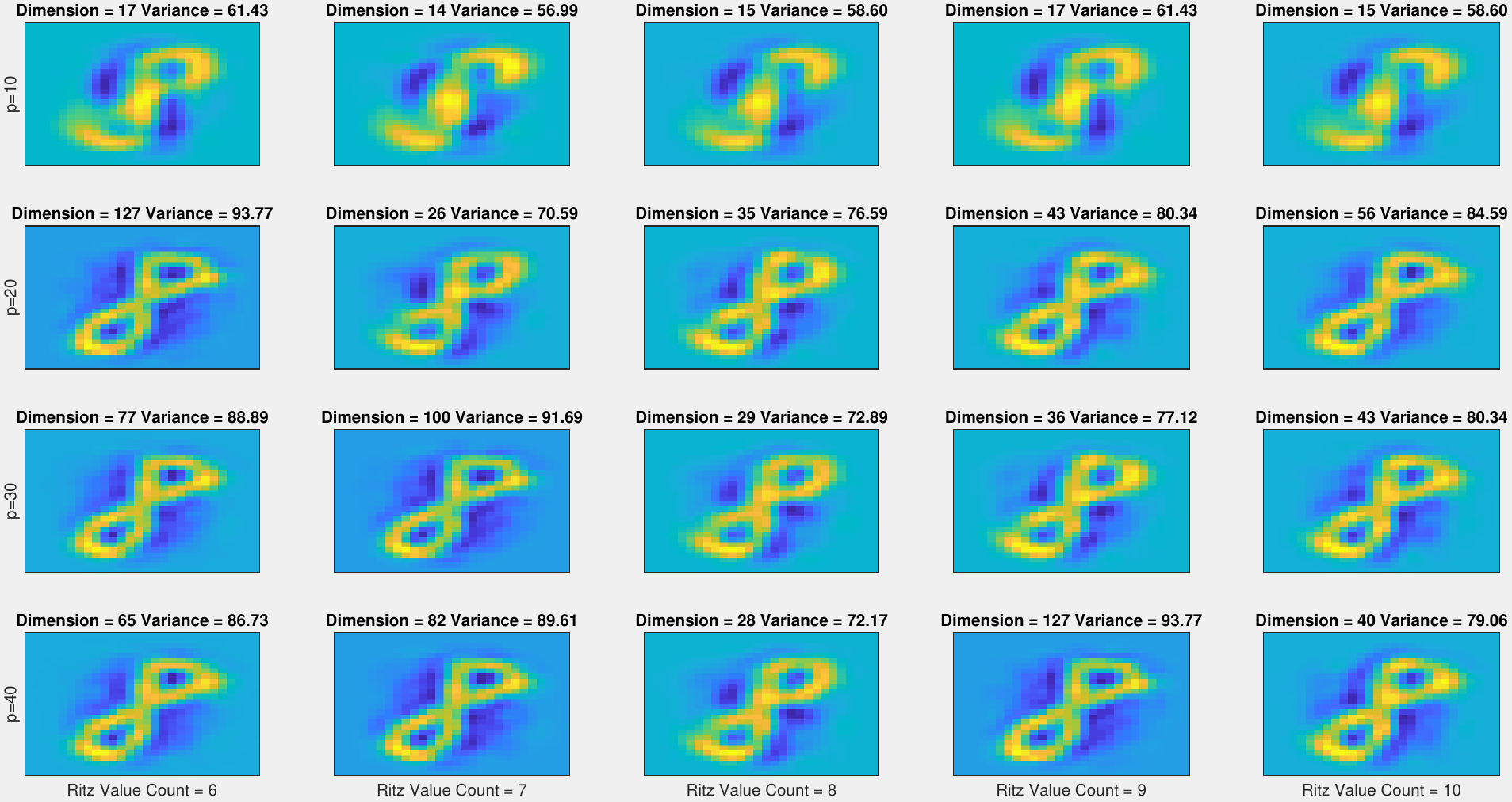}
\subsection{Comparisons with other estimators}

\begin{landscape}
\begin{table}[h]
\caption{Comprasion of different ID estimators}\label{tab1}%

Based on~\citep{scikit-dimension}.
\begin{adjustbox}{scale = 0.65}
\begin{tabular}{|c|c|c|c|c|c|c|c|c|c|c|c|c|c|c|c|c|c|c|c|c|c|c|c|c|}
\toprule
    &ID & OM & OMke & OMk4  & OMse & OMs4 & PF & PFN & PMG & PRT & PPT & PKS & PBS & CID & FSH & KNN & MDA & MND & MLE & MOM & TLE & TNN & DNC & ESS \\
\midrule
 M1Sphere & 10.00 & 11.00 & 7.34 & 8.00 & 7.05 & 7.75 & 11.00 & 7.00 & 1.00 & 9.00 & 10.94 & 5.00 & 1.00 & 8.91 & 11.02 & 11.00 & 9.63 & 9.10 & 9.53 & 8.60 & 10.10 & 9.28 & 11.00 & 10.18  \\  
M2Affine3to5 & 3.00 & 2.00 & 4.36 & 4.25 & 3.73 & 3.50 & 3.00 & 2.00 & 3.00 & 2.00 & 2.27 & 2.00 & 3.00 & 2.87 & 2.92 & 1.00 & 3.07 & 2.90 & 3.03 & 2.79 & 3.13 & 2.92 & 2.97 & 2.90  \\  
M3Nonlinear4to6 & 4.00 & 2.00 & 3.17 & 3.25 & 2.90 & 3.25 & 5.00 & 2.00 & 2.00 & 3.00 & 2.75 & 2.00 & 3.00 & 3.58 & 3.34 & 2.00 & 4.14 & 3.77 & 4.03 & 3.77 & 4.21 & 3.79 & 5.27 & 4.02  \\  
M4Nonlinear & 4.00 & 6.00 & 5.56 & 4.25 & 5.36 & 6.00 & 8.00 & 5.00 & 1.00 & 7.00 & 7.96 & 4.00 & 1.00 & 3.75 & 5.81 & 2.00 & 4.34 & 3.86 & 4.18 & 4.49 & 4.42 & 3.74 & 5.21 & 5.03  \\  
M5aHelix1d & 1.00 & 3.00 & 1.35 & 2.75 & 0.00 & 0.00 & 3.00 & 2.00 & 2.00 & 2.00 & 2.40 & 2.00 & 1.00 & 1.01 & 2.91 & 1.00 & 1.08 & 1.00 & 1.06 & 1.17 & 1.11 & 0.98 & 1.00 & 1.63  \\  
M5bHelix2d & 2.00 & 3.00 & 2.86 & 3.00 & 0.00 & 2.25 & 3.00 & 2.00 & 2.00 & 2.00 & 2.23 & 2.00 & 1.00 & 2.79 & 2.91 & 1.00 & 3.17 & 2.37 & 2.82 & 2.86 & 2.91 & 1.97 & 2.47 & 2.88  \\  
M6Nonlinear & 6.00 & 9.00 & 8.90 & 7.00 & 21.52 & 7.50 & 12.00 & 9.00 & 12.00 & 10.00 & 11.95 & 12.00 & 1.00 & 6.00 & 8.43 & 36.00 & 7.34 & 6.25 & 6.91 & 7.16 & 7.12 & 5.87 & 8.61 & 8.52  \\  
M7Roll & 2.00 & 3.00 & 2.81 & 2.50 & 0.00 & 3.00 & 3.00 & 2.00 & 1.00 & 3.00 & 2.95 & 1.00 & 1.00 & 1.96 & 2.88 & 3.00 & 2.09 & 2.00 & 2.07 & 2.01 & 2.14 & 1.94 & 2.22 & 2.15  \\  
M8Nonlinear & 12.00 & 18.00 & 13.19 & 16.00 & 28.60 & 12.75 & 24.00 & 17.00 & 24.00 & 19.00 & 23.77 & 24.00 & 1.00 & 11.33 & 17.95 & 6.00 & 14.43 & 10.00 & 14.24 & 13.21 & 13.83 & 13.80 & 11.99 & 19.50  \\  
M9Affine & 20.00 & 15.00 & 11.18 & 13.75 & 12.14 & 14.25 & 20.00 & 11.00 & 19.00 & 16.00 & 19.86 & 9.00 & 1.00 & 13.48 & 19.07 & 7.00 & 15.12 & 10.00 & 15.14 & 13.49 & 15.25 & 14.77 & 19.00 & 19.41  \\  
M10aCubic & 10.00 & 11.00 & 7.12 & 8.00 & 6.71 & 7.50 & 11.00 & 6.00 & 2.00 & 9.00 & 10.97 & 5.00 & 1.00 & 8.58 & 10.43 & 6.00 & 9.15 & 8.89 & 9.20 & 8.39 & 9.62 & 9.10 & 11.00 & 10.30  \\  
M10bCubic & 17.00 & 14.00 & 10.30 & 12.50 & 10.47 & 13.25 & 18.00 & 10.00 & 17.00 & 14.00 & 17.88 & 8.00 & 1.00 & 12.81 & 17.10 & 10.00 & 13.96 & 10.00 & 13.90 & 12.47 & 14.20 & 13.96 & 18.00 & 17.29  \\  
M10cCubic & 24.00 & 19.00 & 12.80 & 17.75 & 12.05 & 17.25 & 25.00 & 14.00 & 23.00 & 20.00 & 24.76 & 14.00 & 1.00 & 16.41 & 23.27 & 5.00 & 17.87 & 10.00 & 17.97 & 15.90 & 18.01 & 18.05 & 25.00 & 24.38  \\  
M10dCubic & 70.00 & 52.00 & 23.36 & 48.50 & 22.27 & 45.75 & 71.00 & 39.00 & 69.00 & 53.00 & 69.05 & 35.00 & 1.00 & 31.62 & NaN & 71.00 & 36.65 & 10.00 & 37.17 & 31.47 & 34.71 & 39.85 & 71.00 & 69.95  \\  
M11Moebius & 2.00 & 3.00 & 2.70 & 3.00 & 2.61 & 3.00 & 3.00 & 1.00 & 2.00 & 2.00 & 2.15 & 2.00 & 1.00 & 1.97 & 1.97 & 3.00 & 2.10 & 2.00 & 2.07 & 2.21 & 2.16 & 2.02 & 2.25 & 2.49  \\  
M12Norm & 20.00 & 15.00 & 11.51 & 14.00 & 11.60 & 13.75 & 20.00 & 11.00 & 11.00 & 16.00 & 19.82 & 11.00 & 1.00 & 12.74 & 20.30 & 19.00 & 16.04 & 10.00 & 16.22 & 13.98 & 15.47 & 17.04 & 12.00 & 19.79  \\  
M13aScurve & 2.00 & 3.00 & 0.00 & 2.75 & 0.00 & 3.00 & 3.00 & 1.00 & 1.00 & 2.00 & 1.86 & 1.00 & 2.00 & 1.94 & 2.88 & 3.00 & 2.08 & 2.00 & 2.06 & 1.93 & 2.13 & 1.96 & 2.22 & 2.05  \\  
M13bSpiral & 1.00 & 13.00 & 0.00 & 0.00 & 0.00 & 0.00 & 2.00 & 1.00 & 3.00 & 2.00 & 2.03 & 2.00 & 3.00 & 4.01 & 1.97 & 1.00 & 3.78 & 1.20 & 1.75 & 1.83 & 1.84 & 1.00 & 1.00 & 1.95  \\  
Mbeta & 10.00 & 8.00 & 14.95 & 22.50 & 16.05 & 22.75 & 10.00 & 7.00 & 20.00 & 9.00 & 10.62 & 10.00 & 11.00 & 3.47 & 5.27 & 2.00 & 6.64 & 6.31 & 6.63 & 5.78 & 6.21 & 6.54 & 8.01 & 5.92  \\  
Mn1Nonlinear & 18.00 & 14.00 & 12.48 & 13.75 & 18.17 & 13.50 & 27.00 & 13.00 & 36.00 & 15.00 & 18.82 & 18.00 & 19.00 & 12.58 & 17.00 & 3.00 & 14.31 & 10.00 & 14.26 & 12.71 & 14.29 & 14.24 & 21.00 & 18.39  \\  
Mn2Nonlinear & 24.00 & 18.00 & 12.67 & 17.75 & 19.36 & 17.75 & 36.00 & 18.00 & 48.00 & 19.00 & 25.08 & 24.00 & 25.00 & 15.42 & 23.30 & 3.00 & 17.87 & 10.00 & 17.85 & 15.63 & 17.49 & 18.26 & 29.00 & 24.85  \\  
Mp1Paraboloid & 3.00 & 2.00 & 6.38 & 6.50 & 4.52 & 11.25 & 1.00 & 1.00 & 11.00 & 1.00 & 1.24 & 1.00 & 2.00 & 2.14 & 0.90 & 12.00 & 3.07 & 2.90 & 3.04 & 2.73 & 3.13 & 2.99 & 3.14 & 2.85  \\  
Mp2Paraboloid & 6.00 & 2.00 & 10.27 & 5.75 & 8.50 & 1.50 & 1.00 & 21.00 & 20.00 & 1.00 & 1.06 & 1.00 & 2.00 & 2.63 & 0.89 & 1.00 & 5.14 & 5.08 & 5.19 & 4.22 & 5.34 & 5.48 & 7.02 & 4.89  \\  
Mp3Paraboloid & 9.00 & 2.00 & 6.67 & 1.25 & 11.09 & 8.75 & 1.00 & 1.00 & 29.00 & 1.00 & 1.03 & 1.00 & 2.00 & 2.92 & 0.89 & 4.00 & 6.55 & 6.52 & 6.64 & 4.76 & 6.70 & 7.23 & 8.96 & 6.12  \\

\botrule
\end{tabular}
\end{adjustbox}

\end{table}
\end{landscape}

\begin{table}[h]
\caption{Comprasion of Our Method}\label{tab1}%
\begin{tabular}{|c|c|c|c|c|c|c|}
\toprule
    &ID & OM & OMke & OMk4  & OMse & OMs4 \\
\midrule
 M1Sphere & 10.00 & 11.00 & 7.34 & 8.00 & 7.05 & 7.75  \\  
M2Affine3to5 & 3.00 & 2.00 & 4.36 & 4.25 & 3.73 & 3.50  \\  
M3Nonlinear4to6 & 4.00 & 2.00 & 3.17 & 3.25 & 2.90 & 3.25  \\  
M4Nonlinear & 4.00 & 6.00 & 5.56 & 4.25 & 5.36 & 6.00  \\  
M5aHelix1d & 1.00 & 3.00 & 1.35 & 2.75 & 0.00 & 0.00  \\  
M5bHelix2d & 2.00 & 3.00 & 2.86 & 3.00 & 0.00 & 2.25  \\  
M6Nonlinear & 6.00 & 9.00 & 8.90 & 7.00 & 21.52 & 7.50  \\  
M7Roll & 2.00 & 3.00 & 2.81 & 2.50 & 0.00 & 3.00  \\  
M8Nonlinear & 12.00 & 18.00 & 13.19 & 16.00 & 28.60 & 12.75  \\  
M9Affine & 20.00 & 15.00 & 11.18 & 13.75 & 12.14 & 14.25  \\  
M10aCubic & 10.00 & 11.00 & 7.12 & 8.00 & 6.71 & 7.50  \\  
M10bCubic & 17.00 & 14.00 & 10.30 & 12.50 & 10.47 & 13.25  \\  
M10cCubic & 24.00 & 19.00 & 12.80 & 17.75 & 12.05 & 17.25  \\  
M10dCubic & 70.00 & 52.00 & 23.36 & 48.50 & 22.27 & 45.75  \\  
M11Moebius & 2.00 & 3.00 & 2.70 & 3.00 & 2.61 & 3.00  \\  
M12Norm & 20.00 & 15.00 & 11.51 & 14.00 & 11.60 & 13.75  \\  
M13aScurve & 2.00 & 3.00 & 0.00 & 2.75 & 0.00 & 3.00  \\  
M13bSpiral & 1.00 & 13.00 & 0.00 & 0.00 & 0.00 & 0.00  \\  
Mbeta & 10.00 & 8.00 & 14.95 & 22.50 & 16.05 & 22.75  \\  
Mn1Nonlinear & 18.00 & 14.00 & 12.48 & 13.75 & 18.17 & 13.50  \\  
Mn2Nonlinear & 24.00 & 18.00 & 12.67 & 17.75 & 19.36 & 17.75  \\  
Mp1Paraboloid & 3.00 & 2.00 & 6.38 & 6.50 & 4.52 & 11.25  \\  
Mp2Paraboloid & 6.00 & 2.00 & 10.27 & 5.75 & 8.50 & 1.50  \\  
Mp3Paraboloid & 9.00 & 2.00 & 6.67 & 1.25 & 11.09 & 8.75  \\

\botrule
\end{tabular}

\end{table}

\begin{table}[h]
\caption{Comprasion of Our Methods Error}\label{tab1}%
\begin{tabular}{|c c|c c|}
\toprule
    Name & Error & Name & Error  \\
\midrule
TNN & 0.12 & FSH & 0.38 \\  
DNC & 0.13 & PRT & 0.38 \\  
MLE & 0.18 & PPT & 0.39 \\  
ESS & 0.20 & PKS & 0.49 \\  
TLE & 0.20 & PFN & 0.49 \\  
MND & 0.23 & PCA & 0.50 \\  
MOM & 0.25 & PF & 0.50 \\  
MDA & 0.29 & PBS & 0.62 \\  
CID & 0.36 & KNN & 0.77 \\  
 &  & PMG & 0.82  \\

\botrule
\end{tabular}

\end{table}

\begin{table}[h]
\caption{Spectral Clustering Size 2}\label{tab1}%
\begin{tabular}{|c|c|c|c|}
\toprule
    Data Set  & Error & EstID & ID  \\
\midrule
M3Nonlinear4to6 & 0.00 & 4.00 & 4.00 \\  
M4Nonlinear & 0.00 & 4.00 & 4.00 \\  
M11Moebius & 0.00 & 2.00 & 2.00 \\  
M9Affine & 0.03 & 19.50 & 20.00 \\  
M1Sphere & 0.15 & 8.50 & 10.00 \\  
M10aCubic & 0.15 & 8.50 & 10.00 \\  
M8Nonlinear & 0.21 & 9.50 & 12.00 \\  
M10bCubic & 0.24 & 13.00 & 17.00 \\  
M7Roll & 0.25 & 2.50 & 2.00 \\  
M10cCubic & 0.25 & 18.00 & 24.00 \\  
M13aScurve & 0.25 & 2.50 & 2.00 \\  
Mn2Nonlinear & 0.25 & 18.00 & 24.00 \\  
Mn1Nonlinear & 0.28 & 13.00 & 18.00 \\  
M10dCubic & 0.30 & 49.00 & 70.00 \\  
M2Affine3to5 & 0.33 & 2.00 & 3.00 \\  
M6Nonlinear & 0.33 & 8.00 & 6.00 \\  
M12Norm & 0.35 & 13.00 & 20.00 \\  
Mbeta & 0.35 & 6.50 & 10.00 \\  
M5bHelix2d & 0.50 & 3.00 & 2.00 \\  
M13bSpiral & 0.50 & 1.50 & 1.00 \\  
Mp1Paraboloid & 0.50 & 1.50 & 3.00 \\  
Mp2Paraboloid & 0.67 & 2.00 & 6.00 \\  
Mp3Paraboloid & 0.78 & 2.00 & 9.00 \\  
M5aHelix1d & 1.50 & 2.50 & 1.00 \\  
\botrule
\end{tabular}
\end{table}

\begin{table}[h]
\caption{Kmeans Clustering Size 2}\label{tab1}%
\begin{tabular}{|c|c|c|c|}
\toprule
    Data Set  & Error & EstID & ID  \\
\midrule
M5bHelix2d & 0.00 & 2.00 & 2.00 \\  
M13aScurve & 0.00 & 2.00 & 2.00 \\  
M1Sphere & 0.15 & 8.50 & 10.00 \\  
M10aCubic & 0.15 & 8.50 & 10.00 \\  
M2Affine3to5 & 0.17 & 3.50 & 3.00 \\  
Mbeta & 0.20 & 8.00 & 10.00 \\  
M10bCubic & 0.21 & 13.50 & 17.00 \\  
M3Nonlinear4to6 & 0.25 & 3.00 & 4.00 \\  
M10cCubic & 0.25 & 18.00 & 24.00 \\  
Mn1Nonlinear & 0.25 & 13.50 & 18.00 \\  
Mn2Nonlinear & 0.25 & 18.00 & 24.00 \\  
M9Affine & 0.28 & 14.50 & 20.00 \\  
M12Norm & 0.28 & 14.50 & 20.00 \\  
M10dCubic & 0.29 & 49.50 & 70.00 \\  
M6Nonlinear & 0.33 & 8.00 & 6.00 \\  
M8Nonlinear & 0.42 & 17.00 & 12.00 \\  
M4Nonlinear & 0.50 & 6.00 & 4.00 \\  
M7Roll & 0.50 & 3.00 & 2.00 \\  
M11Moebius & 0.50 & 1.00 & 2.00 \\  
Mp1Paraboloid & 0.50 & 1.50 & 3.00 \\  
Mp2Paraboloid & 0.67 & 2.00 & 6.00 \\  
Mp3Paraboloid & 0.78 & 2.00 & 9.00 \\  
M5aHelix1d & 1.00 & 2.00 & 1.00 \\  
M13bSpiral & 5.50 & 6.50 & 1.00 \\  
\botrule
\end{tabular}
\end{table}

\begin{table}[h]
\caption{TwoNN}\label{tab1}%
\begin{tabular}{|c|c|c|c|}
\toprule
    Data Set  & Error & EstID & ID  \\
\midrule
M13bSpiral & 0.00 & 1.00 & 1.00 \\  
Mp1Paraboloid & 0.00 & 2.99 & 3.00 \\  
M11Moebius & 0.01 & 2.02 & 2.00 \\  
M5bHelix2d & 0.01 & 1.97 & 2.00 \\  
M13aScurve & 0.02 & 1.96 & 2.00 \\  
M6Nonlinear & 0.02 & 5.87 & 6.00 \\  
M5aHelix1d & 0.02 & 0.98 & 1.00 \\  
M2Affine3to5 & 0.03 & 2.92 & 3.00 \\  
M7Roll & 0.03 & 1.94 & 2.00 \\  
M3Nonlinear4to6 & 0.05 & 3.79 & 4.00 \\  
M4Nonlinear & 0.06 & 3.74 & 4.00 \\  
M1Sphere & 0.07 & 9.28 & 10.00 \\  
Mp2Paraboloid & 0.09 & 5.48 & 6.00 \\  
M10aCubic & 0.09 & 9.10 & 10.00 \\  
M12Norm & 0.15 & 17.04 & 20.00 \\  
M8Nonlinear & 0.15 & 13.80 & 12.00 \\  
M10bCubic & 0.18 & 13.96 & 17.00 \\  
Mp3Paraboloid & 0.20 & 7.23 & 9.00 \\  
Mn1Nonlinear & 0.21 & 14.24 & 18.00 \\  
Mn2Nonlinear & 0.24 & 18.26 & 24.00 \\  
M10cCubic & 0.25 & 18.05 & 24.00 \\  
M9Affine & 0.26 & 14.77 & 20.00 \\  
Mbeta & 0.35 & 6.54 & 10.00 \\  
M10dCubic & 0.43 & 39.85 & 70.00 \\  

\botrule
\end{tabular}
\end{table}
\newpage

\includegraphics[width=\textwidth]{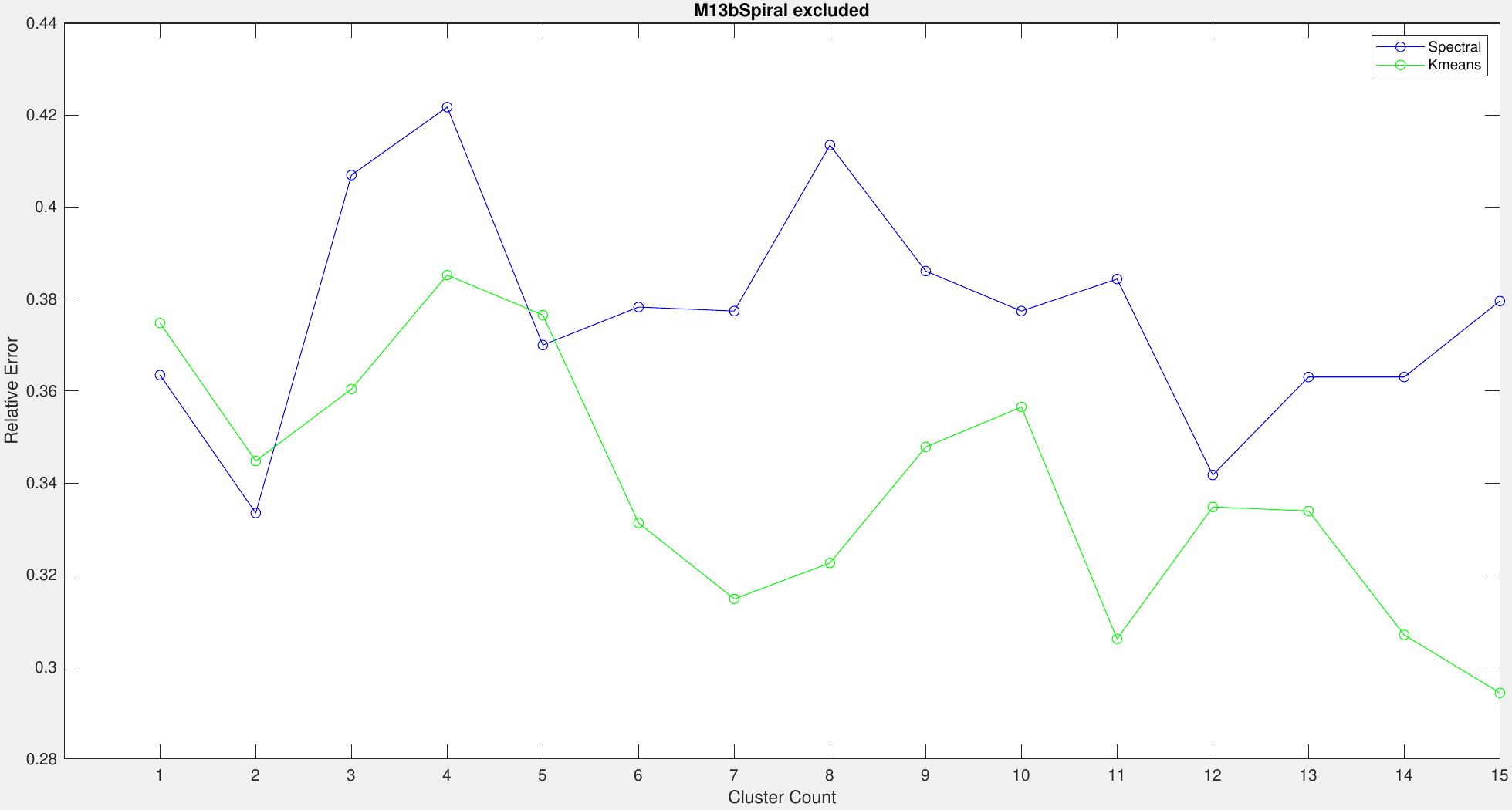}
\includegraphics[width=\textwidth]{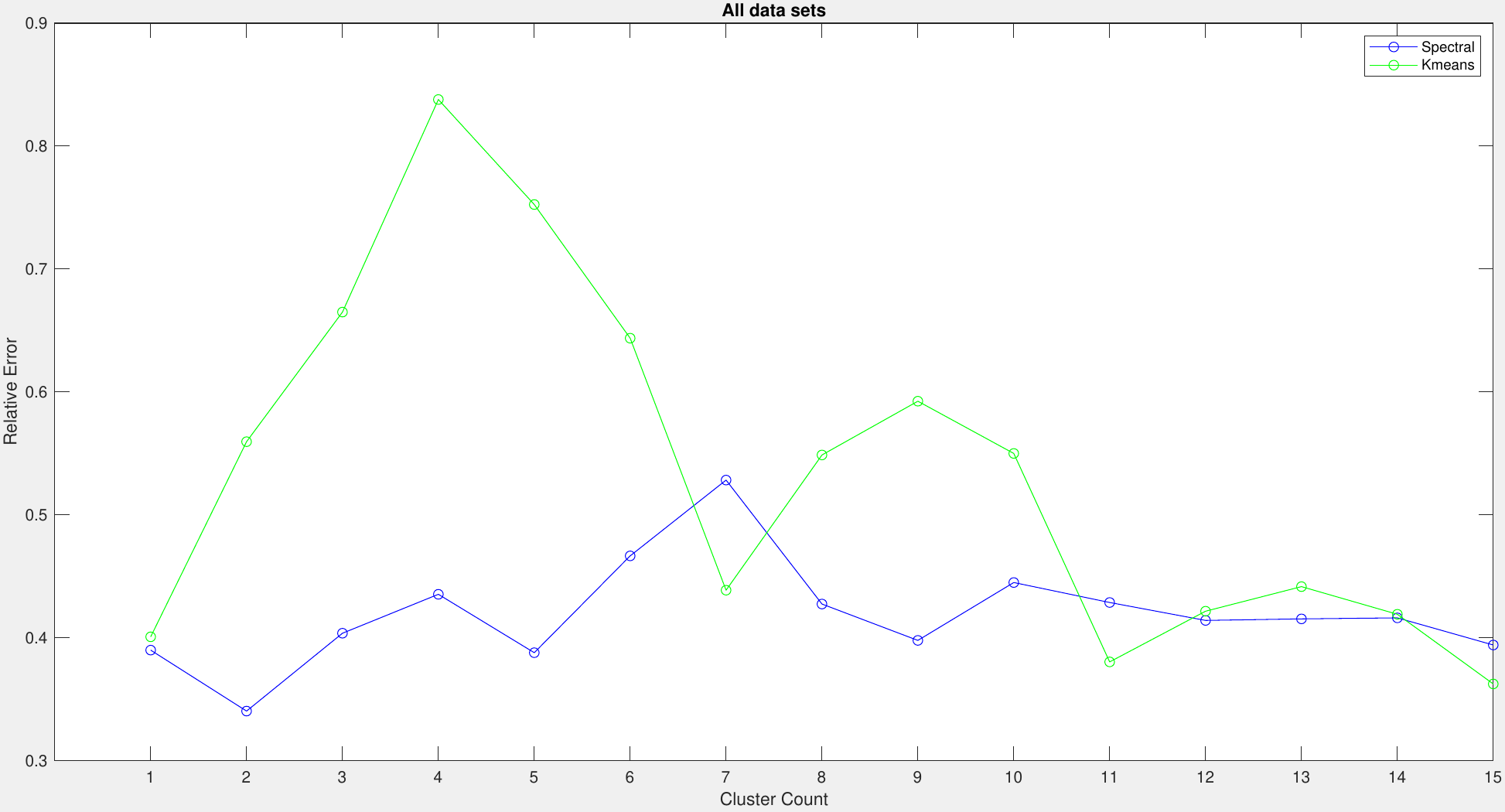}
}

\comment{
\section{Equations}\label{sec4}

Equations in \LaTeX\ can either be inline or on-a-line by itself (``display equations''). For
inline equations use the \verb+$...$+ commands. E.g.: The equation
$H\psi = E \psi$ is written via the command \verb+$H \psi = E \psi$+.

For display equations (with auto generated equation numbers)
one can use the equation or align environments:
\begin{equation}
\|\tilde{X}(k)\|^2 \leq\frac{\sum\limits_{i=1}^{p}\left\|\tilde{Y}_i(k)\right\|^2+\sum\limits_{j=1}^{q}\left\|\tilde{Z}_j(k)\right\|^2 }{p+q}.\label{eq1}
\end{equation}
where,
\begin{align}
D_\mu &=  \partial_\mu - ig \frac{\lambda^a}{2} A^a_\mu \nonumber \\
F^a_{\mu\nu} &= \partial_\mu A^a_\nu - \partial_\nu A^a_\mu + g f^{abc} A^b_\mu A^a_\nu \label{eq2}
\end{align}
Notice the use of \verb+\nonumber+ in the align environment at the end
of each line, except the last, so as not to produce equation numbers on
lines where no equation numbers are required. The \verb+\label{}+ command
should only be used at the last line of an align environment where
\verb+\nonumber+ is not used.
\begin{equation}
Y_\infty = \left( \frac{m}{\textrm{GeV}} \right)^{-3}
    \left[ 1 + \frac{3 \ln(m/\textrm{GeV})}{15}
    + \frac{\ln(c_2/5)}{15} \right]
\end{equation}
The class file also supports the use of \verb+\mathbb{}+, \verb+\mathscr{}+ and
\verb+\mathcal{}+ commands. As such \verb+\mathbb{R}+, \verb+\mathscr{R}+
and \verb+\mathcal{R}+ produces $\mathbb{R}$, $\mathscr{R}$ and $\mathcal{R}$
respectively (refer Subsubsection~\ref{subsubsec2}).

\section{Tables}\label{sec5}

Tables can be inserted via the normal table and tabular environment. To put
footnotes inside tables you should use \verb+\footnotetext[]{...}+ tag.
The footnote appears just below the table itself (refer Tables~\ref{tab1} and \ref{tab2}). 
For the corresponding footnotemark use \verb+\footnotemark[...]+

\begin{table}[h]
\begin{center}
\begin{minipage}{174pt}
\caption{Caption text}\label{tab1}%
\begin{tabular}{@{}llll@{}}
\toprule
Column 1 & Column 2  & Column 3 & Column 4\\
\midrule
row 1    & data 1   & data 2  & data 3  \\
row 2    & data 4   & data 5\footnotemark[1]  & data 6  \\
row 3    & data 7   & data 8  & data 9\footnotemark[2]  \\
\botrule
\end{tabular}
\footnotetext{Source: This is an example of table footnote. This is an example of table footnote.}
\footnotetext[1]{Example for a first table footnote. This is an example of table footnote.}
\footnotetext[2]{Example for a second table footnote. This is an example of table footnote.}
\end{minipage}
\end{center}
\end{table}

\noindent
The input format for the above table is as follows:

\bigskip
\begin{verbatim}
\begin{table}[<placement-specifier>]
\begin{center}
\begin{minipage}{<preferred-table-width>}
\caption{<table-caption>}\label{<table-label>}%
\begin{tabular}{@{}llll@{}}
\toprule
Column 1 & Column 2 & Column 3 & Column 4\\
\midrule
row 1 & data 1 & data 2	 & data 3 \\
row 2 & data 4 & data 5\footnotemark[1] & data 6 \\
row 3 & data 7 & data 8	 & data 9\footnotemark[2]\\
\botrule
\end{tabular}
\footnotetext{Source: This is an example of table footnote. 
This is an example of table footnote.}
\footnotetext[1]{Example for a first table footnote.
This is an example of table footnote.}
\footnotetext[2]{Example for a second table footnote. 
This is an example of table footnote.}
\end{minipage}
\end{center}
\end{table}
\end{verbatim}
\bigskip

\begin{table}[h]
\begin{center}
\begin{minipage}{\textwidth}
\caption{Example of a lengthy table which is set to full textwidth}\label{tab2}
\begin{tabular*}{\textwidth}{@{\extracolsep{\fill}}lcccccc@{\extracolsep{\fill}}}
\toprule%
& \multicolumn{3}{@{}c@{}}{Element 1\footnotemark[1]} & \multicolumn{3}{@{}c@{}}{Element 2\footnotemark[2]} \\\cmidrule{2-4}\cmidrule{5-7}%
Project & Energy & $\sigma_{calc}$ & $\sigma_{expt}$ & Energy & $\sigma_{calc}$ & $\sigma_{expt}$ \\
\midrule
Element 3  & 990 A & 1168 & $1547\pm12$ & 780 A & 1166 & $1239\pm100$\\
Element 4  & 500 A & 961  & $922\pm10$  & 900 A & 1268 & $1092\pm40$\\
\botrule
\end{tabular*}
\footnotetext{Note: This is an example of table footnote. This is an example of table footnote this is an example of table footnote this is an example of~table footnote this is an example of table footnote.}
\footnotetext[1]{Example for a first table footnote.}
\footnotetext[2]{Example for a second table footnote.}
\end{minipage}
\end{center}
\end{table}

In case of double column layout, tables which do not fit in single column width should be set to full text width. For this, you need to use \verb+\begin{table*}+ \verb+...+ \verb+\end{table*}+ instead of \verb+\begin{table}+ \verb+...+ \verb+\end{table}+ environment. Lengthy tables which do not fit in textwidth should be set as rotated table. For this, you need to use \verb+\begin{sidewaystable}+ \verb+...+ \verb+\end{sidewaystable}+ instead of \verb+\begin{table*}+ \verb+...+ \verb+\end{table*}+ environment. This environment puts tables rotated to single column width. For tables rotated to double column width, use \verb+\begin{sidewaystable*}+ \verb+...+ \verb+\end{sidewaystable*}+.

\begin{sidewaystable}
\sidewaystablefn%
\begin{center}
\begin{minipage}{\textheight}
\caption{Tables which are too long to fit, should be written using the ``sidewaystable'' environment as shown here}\label{tab3}
\begin{tabular*}{\textheight}{@{\extracolsep{\fill}}lcccccc@{\extracolsep{\fill}}}
\toprule%
& \multicolumn{3}{@{}c@{}}{Element 1\footnotemark[1]}& \multicolumn{3}{@{}c@{}}{Element\footnotemark[2]} \\\cmidrule{2-4}\cmidrule{5-7}%
Projectile & Energy	& $\sigma_{calc}$ & $\sigma_{expt}$ & Energy & $\sigma_{calc}$ & $\sigma_{expt}$ \\
\midrule
Element 3 & 990 A & 1168 & $1547\pm12$ & 780 A & 1166 & $1239\pm100$ \\
Element 4 & 500 A & 961  & $922\pm10$  & 900 A & 1268 & $1092\pm40$ \\
Element 5 & 990 A & 1168 & $1547\pm12$ & 780 A & 1166 & $1239\pm100$ \\
Element 6 & 500 A & 961  & $922\pm10$  & 900 A & 1268 & $1092\pm40$ \\
\botrule
\end{tabular*}
\footnotetext{Note: This is an example of table footnote this is an example of table footnote this is an example of table footnote this is an example of~table footnote this is an example of table footnote.}
\footnotetext[1]{This is an example of table footnote.}
\end{minipage}
\end{center}
\end{sidewaystable}

\section{Figures}\label{sec6}

As per the \LaTeX\ standards you need to use eps images for \LaTeX\ compilation and \verb+pdf/jpg/png+ images for \verb+PDFLaTeX+ compilation. This is one of the major difference between \LaTeX\ and \verb+PDFLaTeX+. Each image should be from a single input .eps/vector image file. Avoid using subfigures. The command for inserting images for \LaTeX\ and \verb+PDFLaTeX+ can be generalized. The package used to insert images in \verb+LaTeX/PDFLaTeX+ is the graphicx package. Figures can be inserted via the normal figure environment as shown in the below example:

\bigskip
\begin{verbatim}
\begin{figure}[<placement-specifier>]
\centering
\includegraphics{<eps-file>}
\caption{<figure-caption>}\label{<figure-label>}
\end{figure}
\end{verbatim}
\bigskip

\begin{figure}[h]%
\centering
\includegraphics[width=0.9\textwidth]{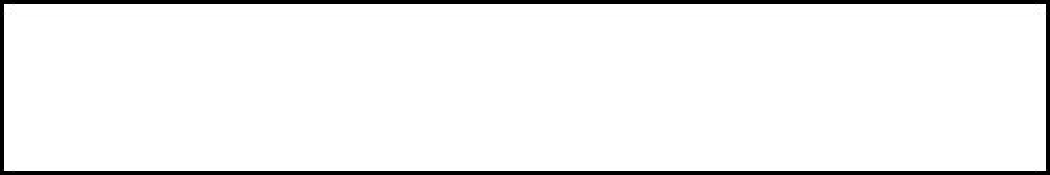}
\caption{This is a widefig. This is an example of long caption this is an example of long caption  this is an example of long caption this is an example of long caption}\label{fig1}
\end{figure}

In case of double column layout, the above format puts figure captions/images to single column width. To get spanned images, we need to provide \verb+\begin{figure*}+ \verb+...+ \verb+\end{figure*}+.

For sample purpose, we have included the width of images in the optional argument of \verb+\includegraphics+ tag. Please ignore this. 

\section{Algorithms, Program codes and Listings}\label{sec7}

Packages \verb+algorithm+, \verb+algorithmicx+ and \verb+algpseudocode+ are used for setting algorithms in \LaTeX\ using the format:

\bigskip
\begin{verbatim}
\begin{algorithm}
\caption{<alg-caption>}\label{<alg-label>}
\begin{algorithmic}[1]
. . .
\end{algorithmic}
\end{algorithm}
\end{verbatim}
\bigskip

You may refer above listed package documentations for more details before setting \verb+algorithm+ environment. For program codes, the ``program'' package is required and the command to be used is \verb+\begin{program}+ \verb+...+ \verb+\end{program}+. A fast exponentiation procedure:

\begin{program}
\BEGIN \\ %
  \FOR i:=1 \TO 10 \STEP 1 \DO
     |expt|(2,i); \\ |newline|() \OD %
\rcomment{Comments will be set flush to the right margin}
\WHERE
\PROC |expt|(x,n) \BODY
          z:=1;
          \DO \IF n=0 \THEN \EXIT \FI;
             \DO \IF |odd|(n) \THEN \EXIT \FI;
\COMMENT{This is a comment statement};
                n:=n/2; x:=x*x \OD;
             \{ n>0 \};
             n:=n-1; z:=z*x \OD;
          |print|(z) \ENDPROC
\END
\end{program}

\begin{algorithm}
\caption{Compute $y = x^n$}\label{algo1}
\begin{algorithmic}[1]
\Require $n \geq 0 \vee x \neq 0$
\Ensure $y = x^n$ 
\State $y \Leftarrow 1$
\If{$n < 0$}\label{algln2}
        \State $X \Leftarrow 1 / x$
        \State $N \Leftarrow -n$
\Else
        \State $X \Leftarrow x$
        \State $N \Leftarrow n$
\EndIf
\While{$N \neq 0$}
        \If{$N$ is even}
            \State $X \Leftarrow X \times X$
            \State $N \Leftarrow N / 2$
        \Else[$N$ is odd]
            \State $y \Leftarrow y \times X$
            \State $N \Leftarrow N - 1$
        \EndIf
\EndWhile
\end{algorithmic}
\end{algorithm}
\bigskip

Similarly, for \verb+listings+, use the \verb+listings+ package. \verb+\begin{lstlisting}+ \verb+...+ \verb+\end{lstlisting}+ is used to set environments similar to \verb+verbatim+ environment. Refer to the \verb+lstlisting+ package documentation for more details.

\bigskip
\begin{minipage}{\hsize}%
\lstset{frame=single,framexleftmargin=-1pt,framexrightmargin=-17pt,framesep=12pt,linewidth=0.98\textwidth,language=pascal}
\begin{lstlisting}
for i:=maxint to 0 do
begin
{ do nothing }
end;
Write('Case insensitive ');
Write('Pascal keywords.');
\end{lstlisting}
\end{minipage}

\section{Cross referencing}\label{sec8}

Environments such as figure, table, equation and align can have a label
declared via the \verb+\label{#label}+ command. For figures and table
environments use the \verb+\label{}+ command inside or just
below the \verb+\caption{}+ command. You can then use the
\verb+\ref{#label}+ command to cross-reference them. As an example, consider
the label declared for Figure~\ref{fig1} which is
\verb+\label{fig1}+. To cross-reference it, use the command 
\verb+Figure \ref{fig1}+, for which it comes up as
``Figure~\ref{fig1}''. 

To reference line numbers in an algorithm, consider the label declared for the line number 2 of Algorithm~\ref{algo1} is \verb+\label{algln2}+. To cross-reference it, use the command \verb+\ref{algln2}+ for which it comes up as line~\ref{algln2} of Algorithm~\ref{algo1}.

\subsection{Details on reference citations}\label{subsec7}

Standard \LaTeX\ permits only numerical citations. To support both numerical and author-year citations this template uses \verb+natbib+ \LaTeX\ package. For style guidance please refer to the template user manual.

Here is an example for \verb+\citep{...}+: \citep{bib1}. Another example for \verb+\citepp{...}+: \citepp{bib2}. For author-year citation mode, \verb+\citep{...}+ prints Jones et al. (1990) and \verb+\citepp{...}+ prints (Jones et al., 1990).

All cited bib entries are printed at the end of this article: \citep{bib3}, \citep{bib4}, \citep{bib5}, \citep{bib6}, \citep{bib7}, \citep{bib8}, \citep{bib9}, \citep{bib10}, \citep{bib11} and \citep{bib12}.

\section{Examples for theorem like environments}\label{sec10}

For theorem like environments, we require \verb+amsthm+ package. There are three types of predefined theorem styles exists---\verb+thmstyleone+, \verb+thmstyletwo+ and \verb+thmstylethree+ 

\bigskip
\begin{tabular}{|l|p{19pc}|}
\hline
\verb+thmstyleone+ & Numbered, theorem head in bold font and theorem text in italic style \\\hline
\verb+thmstyletwo+ & Numbered, theorem head in roman font and theorem text in italic style \\\hline
\verb+thmstylethree+ & Numbered, theorem head in bold font and theorem text in roman style \\\hline
\end{tabular}
\bigskip

For mathematics journals, theorem styles can be included as shown in the following examples:

\begin{theorem}[Theorem subhead]\label{thm1}
Example theorem text. Example theorem text. Example theorem text. Example theorem text. Example theorem text. 
Example theorem text. Example theorem text. Example theorem text. Example theorem text. Example theorem text. 
Example theorem text. 
\end{theorem}

Sample body text. Sample body text. Sample body text. Sample body text. Sample body text. Sample body text. Sample body text. Sample body text.

\begin{proposition}
Example proposition text. Example proposition text. Example proposition text. Example proposition text. Example proposition text. 
Example proposition text. Example proposition text. Example proposition text. Example proposition text. Example proposition text. 
\end{proposition}

Sample body text. Sample body text. Sample body text. Sample body text. Sample body text. Sample body text. Sample body text. Sample body text.

\begin{example}
Phasellus adipiscing semper elit. Proin fermentum massa
ac quam. Sed diam turpis, molestie vitae, placerat a, molestie nec, leo. Maecenas lacinia. Nam ipsum ligula, eleifend
at, accumsan nec, suscipit a, ipsum. Morbi blandit ligula feugiat magna. Nunc eleifend consequat lorem. 
\end{example}

Sample body text. Sample body text. Sample body text. Sample body text. Sample body text. Sample body text. Sample body text. Sample body text.

\begin{remark}
Phasellus adipiscing semper elit. Proin fermentum massa
ac quam. Sed diam turpis, molestie vitae, placerat a, molestie nec, leo. Maecenas lacinia. Nam ipsum ligula, eleifend
at, accumsan nec, suscipit a, ipsum. Morbi blandit ligula feugiat magna. Nunc eleifend consequat lorem. 
\end{remark}

Sample body text. Sample body text. Sample body text. Sample body text. Sample body text. Sample body text. Sample body text. Sample body text.

\begin{definition}[Definition sub head]
Example definition text. Example definition text. Example definition text. Example definition text. Example definition text. Example definition text. Example definition text. Example definition text. 
\end{definition}

Additionally a predefined ``proof'' environment is available: \verb+\begin{proof}+ \verb+...+ \verb+\end{proof}+. This prints a ``Proof'' head in italic font style and the ``body text'' in roman font style with an open square at the end of each proof environment. 

\begin{proof}
Example for proof text. Example for proof text. Example for proof text. Example for proof text. Example for proof text. Example for proof text. Example for proof text. Example for proof text. Example for proof text. Example for proof text. 
\end{proof}

Sample body text. Sample body text. Sample body text. Sample body text. Sample body text. Sample body text. Sample body text. Sample body text.

\begin{proof}[Proof of Theorem~{\upshape\ref{thm1}}]
Example for proof text. Example for proof text. Example for proof text. Example for proof text. Example for proof text. Example for proof text. Example for proof text. Example for proof text. Example for proof text. Example for proof text. 
\end{proof}

\noindent
For a quote environment, use \verb+\begin{quote}...\end{quote}+
\begin{quote}
Quoted text example. Aliquam porttitor quam a lacus. Praesent vel arcu ut tortor cursus volutpat. In vitae pede quis diam bibendum placerat. Fusce elementum
convallis neque. Sed dolor orci, scelerisque ac, dapibus nec, ultricies ut, mi. Duis nec dui quis leo sagittis commodo.
\end{quote}

Sample body text. Sample body text. Sample body text. Sample body text. Sample body text (refer Figure~\ref{fig1}). Sample body text. Sample body text. Sample body text (refer Table~\ref{tab3}). 

\section{Methods}\label{sec11}

Topical subheadings are allowed. Authors must ensure that their Methods section includes adequate experimental and characterization data necessary for others in the field to reproduce their work. Authors are encouraged to include RIIDs where appropriate. 

\textbf{Ethical approval declarations} (only required where applicable) Any article reporting experiment/s carried out on (i)~live vertebrate (or higher invertebrates), (ii)~humans or (iii)~human samples must include an unambiguous statement within the methods section that meets the following requirements: 

\begin{enumerate}[1.]
\item Approval: a statement which confirms that all experimental protocols were approved by a named institutional and/or licensing committee. Please identify the approving body in the methods section

\item Accordance: a statement explicitly saying that the methods were carried out in accordance with the relevant guidelines and regulations

\item Informed consent (for experiments involving humans or human tissue samples): include a statement confirming that informed consent was obtained from all participants and/or their legal guardian/s
\end{enumerate}

If your manuscript includes potentially identifying patient/participant information, or if it describes human transplantation research, or if it reports results of a clinical trial then  additional information will be required. Please visit (\url{https://www.nature.com/nature-research/editorial-policies}) for Nature Portfolio journals, (\url{https://www.springer.com/gp/authors-editors/journal-author/journal-author-helpdesk/publishing-ethics/14214}) for Springer Nature journals, or (\url{https://www.biomedcentral.com/getpublished/editorial-policies\#ethics+and+consent}) for BMC.

\section{Discussion}\label{sec12}

Discussions should be brief and focused. In some disciplines use of Discussion or `Conclusion' is interchangeable. It is not mandatory to use both. Some journals prefer a section `Results and Discussion' followed by a section `Conclusion'. Please refer to Journal-level guidance for any specific requirements. 

\section{Conclusion}\label{sec13}

Conclusions may be used to restate your hypothesis or research question, restate your major findings, explain the relevance and the added value of your work, highlight any limitations of your study, describe future directions for research and recommendations. 

In some disciplines use of Discussion or 'Conclusion' is interchangeable. It is not mandatory to use both. Please refer to Journal-level guidance for any specific requirements. 

\backmatter

\bmhead{Supplementary information}

If your article has accompanying supplementary file/s please state so here. 

Authors reporting data from electrophoretic gels and blots should supply the full unprocessed scans for key as part of their Supplementary information. This may be requested by the editorial team/s if it is missing.

Please refer to Journal-level guidance for any specific requirements.

\bmhead{Acknowledgments}

Acknowledgments are not compulsory. Where included they should be brief. Grant or contribution numbers may be acknowledged.

Please refer to Journal-level guidance for any specific requirements.

\section*{Declarations}

Some journals require declarations to be submitted in a standardised format. Please check the Instructions for Authors of the journal to which you are submitting to see if you need to complete this section. If yes, your manuscript must contain the following sections under the heading `Declarations':

\begin{itemize}
\item Funding
\item Conflict of interest/Competing interests (check journal-specific guidelines for which heading to use)
\item Ethics approval 
\item Consent to participate
\item Consent for publication
\item Availability of data and materials
\item Code availability 
\item Authors' contributions
\end{itemize}

\noindent
If any of the sections are not relevant to your manuscript, please include the heading and write `Not applicable' for that section. 

\bigskip
\begin{flushleft}%
Editorial Policies for:

\bigskip\noindent
Springer journals and proceedings: \url{https://www.springer.com/gp/editorial-policies}

\bigskip\noindent
Nature Portfolio journals: \url{https://www.nature.com/nature-research/editorial-policies}

\bigskip\noindent
\textit{Scientific Reports}: \url{https://www.nature.com/srep/journal-policies/editorial-policies}

\bigskip\noindent
BMC journals: \url{https://www.biomedcentral.com/getpublished/editorial-policies}
\end{flushleft}

\begin{appendices}

\section{Section title of first appendix}\label{secA1}

An appendix contains supplementary information that is not an essential part of the text itself but which may be helpful in providing a more comprehensive understanding of the research problem or it is information that is too cumbersome to be included in the body of the paper.




\end{appendices}

}

\bibliography{sn-bibliography}


\end{document}